\documentclass[sn-mathphys,Numbered]{sn-jnl}
\pdfoutput=1
\usepackage{graphicx}%
\usepackage{multirow}%
\usepackage{amsmath,amssymb,amsfonts}%
\usepackage{amsthm}%
\usepackage{mathrsfs}%
\usepackage[title]{appendix}%
\usepackage{xcolor}%
\usepackage{textcomp}%
\usepackage{manyfoot}%
\usepackage{booktabs}%
\usepackage{algorithm}%
\usepackage{algorithmicx}%
\usepackage{algpseudocode}%
\usepackage{listings}%
\usepackage{thm-restate}
\usepackage{caption}
\usepackage{subcaption}
\usepackage{comment}

\newcommand{\norm}[1] { \left\Vert #1 \right\Vert }
\theoremstyle{definition}
\newtheorem{definition}{Definition}[section]

\DeclareMathOperator*{\argmin}{arg\,min}

\begin{document}

\title[Advancing Personalized Federated Learning: Group Privacy, Fairness, and Beyond]{Advancing Personalized Federated Learning: Group Privacy, Fairness, and Beyond}


\author*[1,3]{\fnm{Filippo} \sur{Galli}}\email{filippo.galli@sns.it}
\author*[2]{\fnm{Kangsoo} \sur{Jung}}\email{gangsoo.zeong@inria.fr}
\author*[2,4]{\fnm{Sayan} \sur{Biswas}}\email{sayan.biswas@inria.fr}
\author*[2,4]{\fnm{Catuscia} \sur{Palamidessi}} \email{catuscia@lix.polytechnique.fr}
\author*[3]{\fnm{Tommaso} \sur{Cucinotta}}\email{tommaso.cucinotta@santannapisa.it}
\affil[1]{\orgname{Scuola Normale Superiore}, \orgaddress{\city{Pisa},
\country{Italy}}}
\affil[2]{\orgname{INRIA}, \orgaddress{\city{Palaiseau}, \country{France}}}
\affil[3]{\orgname{Scuola Superiore Sant'Anna}, \orgaddress{\city{{Pisa}}, \country{Italy}}}
\affil[4]{\orgname{\'{E}cole Polytechnique}, \orgaddress{\city{{Palaiseau}}, \country{France}}}


\abstract{Federated learning (FL) is a framework for training machine learning models in a distributed and collaborative manner. During training, a set of participating clients process their data stored locally, sharing only updates of the statistical model's parameters obtained by minimizing a cost function over their local inputs. 
FL was proposed as a stepping-stone towards privacy-preserving machine learning, but it has been shown to expose clients to issues such as leakage of private information, lack of personalization of the model, and the possibility of having a trained model that is fairer to some groups of clients than to others. In this paper, the focus is on addressing the triadic interaction among personalization, privacy guarantees, and fairness attained by trained models within the FL framework. Differential privacy and its variants have been studied and applied as cutting-edge standards for providing formal privacy guarantees. However, clients in FL often hold very diverse datasets representing heterogeneous communities, making it important to protect their sensitive and personal information while still ensuring that the trained model upholds the aspect of fairness for the users. To attain this objective, a method is put forth that introduces group privacy assurances through the utilization of \emph{d}-privacy (aka metric privacy). \emph{d}-privacy represents a localized form of differential privacy that relies on a metric-oriented obfuscation approach to maintain the original data's topological distribution. This method, besides enabling personalized model training in a federated approach and providing formal privacy guarantees, possesses significantly better group fairness measured under a variety of standard metrics than a global model trained within a classical FL template. Theoretical justifications for the applicability are provided, as well as experimental validation on real-world datasets to illustrate the working of the proposed method.}


\maketitle

\section{Introduction}
\label{sec:introduction}
The widespread collection of user data in modern machine learning has raised concerns regarding privacy violations and the potential disclosure of sensitive personal information~\cite{lemetayer-hal-01420983, nist}. To address these concerns, Federated Learning~\cite{fed-l-0} was introduced as a collaborative machine learning paradigm, where users' devices train a global predictive model without transmitting raw data to a central server. While FL offers promises of preserving user privacy and maintaining model performance, the heterogeneity of data distributions among clients can lead to challenges such as reduced model utility and convergence issues during training. In response, personalized federated learning approaches have emerged, aiming to tailor models to clusters of users with similar data distributions~\cite{ghosh, mansour, sattler}.

Furthermore, it has been demonstrated that avoiding the release of users' raw data alone does not provide sufficient protection against potential privacy violations~\cite{hitaj2017deep, nasr2019comprehensive, deep-leakage}. To address this issue, researchers have explored the application of differential privacy (DP)~\cite{DworkDP1, DworkDP2} to federated learning, providing privacy guarantees for users participating in the optimization process. DP mechanisms introduce randomness in the model updates released by clients, making each user's contribution to the final model probabilistically indistinguishable up to a certain likelihood factor. To bound this factor, the domain of secrets (i.e., the parameter space in FL) is artificially constrained, either to offer central~\cite{adaptive, brendan2018learning} or local DP guarantees~\cite{ldpfl, zhao2020local}. However, constraining the optimization process to a subset of $\mathbb{R}^n$ can have negative effects, such as when the optimal model parameters for a particular cluster of users lie outside such a bounded domain.

To address the challenges of personalization and local privacy protection, this work proposes the adoption of a more general notion of DP called $d$-privacy or metric-based privacy~\cite{broadening} which has been in the spotlight of late mainly in the context of location-privacy~\cite{biswas2023privic, Natasha_optimaldprivacy, atmaca2022privacy}. This concept of privacy does not require a bounded domain and provides guarantees based on the distance between any two points in the parameter space. Therefore, assuming that clients with similar data distributions have similar optimal fitting parameters, $d$-privacy offers strong indistinguishability guarantees. Conversely, privacy guarantees degrade gracefully for clients with significantly different data distributions.

In addition to addressing privacy concerns in personalised FL as was studied in \cite{Galli_Personalised_icissp23}, this work extends the analysis and investigates the impact of the proposed method on fairness aspects in federated model training. As machine learning-based decision systems become more prevalent, it has become apparent that many of these systems exhibit gender and racial biases that disproportionately affect minority populations~\cite{berk2021fairness, chouldechova2017fair}. Therefore, beyond protecting user privacy, it is crucial to explore cutting-edge machine learning algorithms that can potentially mitigate this pervasive lack of fairness among participating clients. However, systems aiming to protect privacy while ensuring fairness often involve a trade-off between the two~\cite{agarwal2022trade}. 
This trade-off arises because privacy protection techniques based on DP tend to minimize the impact of outliers or minorities within the overall dataset. In other words, the application of \textit{d}-privacy, a metric-based generalization of DP, to personalized FL could potentially compromise the fairness of the machine learning model. Building upon~\cite{Galli_Personalised_icissp23}, this paper presents extensive experimental results demonstrating that the use of personalized FL under group privacy guarantees not only significantly improves fairness compared to the classical (non-personalized) FL framework, but it also maintains a relatively small trade-off between privacy and fairness.

In summary, the contributions of this paper are the following: it  extends the work pursued in~\cite{Galli_Personalised_icissp23}
(points 1 and 2) and it investigates the implications of our proposal on the  fairness of the model (point 3):
\begin{enumerate}
    \item A novel algorithm is put forward for collaborative training of machine learning models, leveraging advanced techniques for model personalization and addressing user privacy concerns by formalizing privacy guarantees in terms of $d$-privacy.
    \item This research focuses on studying the Laplace mechanism under Euclidean distance, and providing a closed-form expression for its generalization in $\mathbb{R}^n$, as well as an efficient sampling procedure.
    \item It shows that personalized federated learning under formal privacy guarantees improves group fairness significantly compared to the non-personalized federated learning framework and, hence, establishes that this method enhances the trade-off between privacy and fairness.
\end{enumerate}
The rest of this paper is organized as follows. Section~\ref{section:Background} introduces the relevant foundations of federated learning, differential privacy, and fairness notions. 
Section ~\ref{sec:related} discusses the related works for our research. Section~\ref{section:Algorithm} explains the proposed algorithm for personalized federated learning with group privacy. 
Section~\ref{section:Experiments} illustrates how the proposed method works in terms of privacy and fairness, and Section~\ref{section:Conclusion} provides our concluding remarks.

\section{Background} \label{section:Background}

\subsection{Personalized Federated Learning}
The problem of personalized federated learning falls within the framework of stochastic optimization, and the notation from~\cite{ghosh} is adopted here to determine the set of minimizers $\theta_j^* \in \mathbb{R}^n$ with $j \in \left\{ 1, \dots, k \right\}$ of the cost functions
\begin{equation} \label{erm:1}
F(\theta_j) = \mathbb{E}_{z\sim\mathcal{D}_j} \left[f(\theta_j;z)\right],
\end{equation}
where $\{\mathcal{D}_1,\ldots,\mathcal{D}_k\}$ are the data distributions which cannot be accessed directly but only through a collection of client datasets $Z_c=\left\{z | z \sim \mathcal{D}_j, z \in \mathbb{D} \right\}$ for some $j\in\{1,\ldots,k\}$ with $c \in C = \left\{ 1, \dots, N \right\}$ the set of clients,
and $\mathbb{D}$ a generic domain of data points. $C$ is partitioned in $k$ disjoint sets 
\begin{equation}\label{erm:2}
S_j^* = \{c\in C \mid \forall z \in Z_c, \, z \sim \mathcal{D}_j\} \quad\forall\,j\in \{1,\ldots,k\}
\end{equation}
The mapping $c \rightarrow j$ is
unknown and it is necessary to rely on estimates $S_j$ of the membership of $Z_c$ to compute
the empirical cost functions 
\begin{equation} \label{erm:3}
\begin{split}
&\tilde{F}(\theta_j) = \frac{1}{|S_j|}\sum_{c \in S_j} \tilde{F_c}(\theta_j;Z_c); \\
&\tilde{F_c}(\theta_j;Z_c) = \frac{1}{|Z_c|}\sum_{z_i \in Z_c}f(\theta; z_i)
\end{split}
\end{equation}
The cost function $f \colon \mathbb{R}^n\times \mathbb{D} \mapsto \mathbb{R}_{\geq 0}$ is applied on $z \in \mathbb{D}$, parametrized by the vector $\theta_j \in \mathbb{R}^n$. Thus, the optimization aims to find, $\forall \,j\,\in\{1,\ldots,k\}$,
\begin{equation} \label{erm:5}
\tilde{\theta}_j^* = \argmin_{\theta_j}\tilde{F}(\theta_j)
\end{equation}

\subsection{Privacy}
$d$-privacy, introduced in~\cite{broadening}, extends the concept of differential privacy (DP) to any domain $\mathcal{X}$, which represents the original data space and is equipped with a distance measure $d\colon \mathcal{X}^2\mapsto \mathbb{R}_{\geq 0}$, along with a space of secrets $\mathcal{Y}$. A random mechanism $\mathcal{R}: \mathcal{X} \mapsto \mathcal{Y}$ is considered $\varepsilon$-$d$-private if, for any $x_1,x_2\in \mathcal{X}$ and measurable $S\subseteq \mathcal{Y}$, the inequality in Equation~\eqref{eq:dprivacy} holds:

\begin{equation}\label{eq:dprivacy}
\mathbb{P}\left[\mathcal{R}(x_1) \in S\right] \leq
e^{\varepsilon d(x_1,x_2)} \mathbb{P}\left[\mathcal{R}(x_2) \in S\right]
\end{equation}

It is important to note that when $\mathcal{X}$ corresponds to the domain of databases and $d$ represents the distance based on the Hamming graph of their adjacency relation, Equation~\eqref{eq:dprivacy} aligns with the standard definition of DP in~\cite{DworkDP1,DworkDP2}. However, in this study, $\theta \in \mathbb{R}^n$ is considered as both the domain $\mathcal{X}$ and the space of secrets $\mathcal{Y}$. The primary motivation behind employing $d$-privacy is to preserve the topology of the parameter distributions among clients. Specifically, it aims to ensure that clients with similar model parameters in the non-privatized space $\mathcal{X}$ will communicate approximate model parameters in the privatized space $\mathcal{Y}$, on average.

\subsection{Fairness}~\label{section:BackgroundFairness}
With the recent surge of interest in building ethical ways to train machine learning models, the topic of fairness in machine learning has been in the spotlight and, correspondingly, various metrics and algorithms to quantify and establish fairness in model training have been studied from a variety of perspectives and in different contexts~\cite{verma2018fairness, hanna2009measuring, makhlouf2021applicability}. Most fairness metrics consider the simple case of having a \emph{privileged} group and an \emph{unprivileged} group in the population. Under this assumption, typically one attribute of the dataset is selected as a sensitive attribute (e.g., gender, race, etc.) that defines the privileged and the unprivileged groups. The goal of fairness in machine learning is to ensure fair and non-discriminated results regardless of the membership in a sensitive attribute. The two main notions of fairness considered by the community are individual fairness and group fairness: \emph{Individual fairness}~\cite{dwork2012fairness} claims that similar individuals should be treated similarly, and \emph{group fairness} requires that different demographic subgroups should receive equal treatment with respect to their sensitive attributes. While both notions of fairness are important, this work focus on group fairness because our goal is to analyze and mitigate the potential bias against certain groups (e.g. demographic groups) through personalization techniques. The following metrics are considered for evaluating group fairness as a part of this work.

 In the rest of the paper, $\hat{Y}=1$, $\hat{Y}=0$ is used to represent the positive and negative prediction respectively, and $S=1$, $S=0$ to represent the privileged and unprivileged group. 

The simplest notion of fairness to be proposed was \emph{demographic parity}~\cite{dwork2012fairness}.

\begin{definition}\label{def:demoparity}
\emph{Demographic parity} is achieved by a system when the prediction $\hat{Y}$ of the target label $Y$ is statistically independent of the sensitive attributes $S$, i.e., 
\begin{equation}
\mathbb{P}\left[\hat{Y}=1| S=1\right]=\mathbb{P}\left[\hat{Y}=1| S=0\right]
\end{equation}
\end{definition}

Imposing demographic parity has often a strong negative impact on accuracy, and, consequently, more refined notions were proposed afterwards. In particular, \emph{equalized odds} and 
\emph{equal opportunity}~\cite{hardt2016equality}.

\begin{definition}\label{def:eqaulodd} A system satisfies \emph{equalized odds} 
if its prediction $\hat{Y}$ is conditionally independent of the sensitive attribute $S$ given the target label $Y$,
\begin{equation}
\mathbb{P}\left[\hat{Y}=1|Y=y, S=1\right]=\mathbb{P}\left[\hat{Y}=1|Y=y, S=0\right], \quad y \in \{0, 1\} \end{equation}

\end{definition}

In other words, the notion of equalized odds requires the privileged and unprivileged groups to have equal true positive rates and equal false positive rates.

Equal opportunity is a relaxation of equalized odds, in the sense that  
it only requires equal true positive rates across the groups.

\begin{definition}\label{def:eqaulopp} \emph{Equal opportunity}  is satisfied by a system if its prediction $\hat{Y}$ is conditionally independent of the sensitive attribute $S$ given the target label $Y$
\begin{equation}
    \mathbb{P}\left[\hat{Y}=1|Y=1, S=1\right]=\mathbb{P}\left[\hat{Y}=1|Y=1, S=0\right]
\end{equation}

\end{definition}

\color{black}

In practice, however, it is difficult to obtain perfect equality for any of the aforementioned notions. Hence, typically the aim is to minimize the absolute value of the difference between the privileged and unprivileged groups, rather than requiring this difference to be exactly zero.  For instance, the \emph{demographic parity difference} is defined as
\begin{equation}
\left\lvert\mathbb{P}\left[\hat{Y}=1| S=1\right] - \mathbb{P}\left[\hat{Y}=1| S=0\right]\right\rvert
\end{equation}
and similarly for the \emph{equalized odd difference} and 
\emph{equal opportunity difference}.

\section{Related Works}\label{sec:related}
Federated optimization has demonstrated suboptimal performance when the local datasets consist of samples from non-congruent distributions, resulting in the inability to simultaneously minimize both client-level and global objectives.
In previous studies~\cite{ghosh, mansour, sattler}, researchers examined various meta-algorithms for personalization, but the assertion of preserving user privacy relies solely on clients releasing updated models or model updates, rather than transferring raw data to the server, which can have significant consequences. To address this issue, several works have focused on the privatization of the (federated) optimization algorithm within the framework of DP~\cite{abadi, flcdp, brendan2018learning, adaptive}, which adopt DP to provide defences against an \textit{honest-but-curious} adversary. However, even in this setting, there is no guarantee of protection against sample reconstruction from the local datasets using client updates, as highlighted in~\cite{deep-leakage}.
Various strategies have been explored to offer local privacy guarantees, either through cryptographic approaches~\cite{secure-aggregation} or within the framework of local DP~\cite{ldpfl, cpsgd, huetal}. Specifically, in~\cite{huetal}, the authors tackle the problem of personalized and locally differentially private federated learning, but only for the case of simple convex, $1$-Lipschitz cost functions of the inputs. It is worth noting that this assumption is unrealistic in the majority of machine learning models and excludes many statistical modelling techniques, particularly neural networks.
Finally, some research focused on designing architectures capable of providing private computing environments for remote users~\cite{bonawitz2017practical}, often making use of trusted platform modules, secure processors~\cite{chhabra2010analysis}, or similar mechanisms~\cite{cucinotta2014confidential} improving efficiency by enforcing encryption on network transmissions, rather than memory accesses. For example, the latter work conceptualizes an architecture that could be leveraged to deploy a server that can only reveal the data being processed to clients that instantiated the server. 
It shall be noted, however, that cryptographic guarantees of security are orthogonal to the
privacy notions of differential privacy and its generalizations. To summarize 
and provide context around this work, Table \ref{tab:comp} provides a qualitative 
evaluation of relevant research and how the contributions presented in this
paper fit among them.

\begin{table}[]
\centering
\begin{tabular}{cccccc}
\multicolumn{1}{l}{} & \multicolumn{1}{l}{\cite{brendan2018learning}} & \multicolumn{1}{l}{\cite{huetal}} & \multicolumn{1}{l}{\cite{ldpfl}} & \multicolumn{1}{l}{\cite{Galli_Personalised_icissp23}} & \multicolumn{1}{l}{This Work} \\
\midrule
Central Privacy & \checkmark & \checkmark & \checkmark & \checkmark  & \checkmark \\
\hline
Local Privacy   & $\times$ & \checkmark & \checkmark & \checkmark  & \checkmark \\
\hline
Personalization & $\times$ & \checkmark & $\times$ & \checkmark  & \checkmark \\
\hline
Mild Assumptions on Training & \checkmark & $\times$ & \checkmark & \checkmark  & \checkmark \\
\hline
Fairness analysis & $\times$ & $\times$ & $\times$ & $\times$ & \checkmark \\
\hline
\end{tabular}
\caption{Qualitative comparison with the most relevant prior research on the topic.}
\label{tab:comp}
\end{table}

Of late, a great deal of attention has been devoted to studying and understanding the aspects of fairness in machine learning~\cite{chhabra2021overview, ezzeldin2021fairfed, chu2021fedfair,menon2018cost, wick2019unlocking,agarwal2022trade,mehrabi2021survey}. Most of the research on fairness focuses on developing techniques to mitigate bias in machine learning algorithms. These techniques can be categorized into three main approaches: pre-processing, in-processing, and post-processing. Pre-processing techniques~\cite{biswas2021fair, kamiran2012data} aim to generate a less biased dataset by modifying the values or adjusting the sampling process. In the case of in-processing techniques~\cite{wan2023processing,hashimoto2018fairness}, the objective function is optimized while taking into account discrimination-aware regularizers. Post-processing techniques~\cite{petersen2021post,noriega2019active} involve adjusting the trained model to produce fairer outcomes. However, it is worth noting that the majority of these studies primarily target centralized machine learning models as opposed to FL. 
Furthermore, there is a lack of research exploring the interplay between accuracy and fairness~\cite{menon2018cost, wick2019unlocking} or privacy and fairness~\cite{agarwal2022trade, cummings2019compatibility}. In particular, to the best of our knowledge, disproportionately fewer works have focused on investigating the relationship between privacy and fairness. \cite{agarwal2022trade} formally proved that privacy and fairness can be at odds with each other with non-trivial accuracy. A few recent works on group fairness in FL have emerged~\cite{ezzeldin2021fairfed, chu2021fedfair} but they do not consider the facet of privacy-fairness trade-off.

\section{An Algorithm for Private and Personalized Federated Learning}
\label{section:Algorithm}

Algorithm~\ref{alg:pifca} aims to enable personalized federated learning while ensuring local privacy guarantees to preserve group privacy. In this context, locality refers to the sanitization of client information before it is shared with the server, while group privacy pertains to the notion of indistinguishability within a specific neighbourhood of clients, defined based on a particular distance metric. To clarify our terminology, we provide definitions for \textit{neighbourhood} and \textit{group} as follows:
\begin{definition}\label{def:neighbourhood}
For any model parameterized by $\theta_0,\in,\mathbb{R}^n$, the $r$-\emph{neighbourhood} is defined as the set of points in the parameter space that are within an $L_2$ distance of $r$ or less from $\theta_0$, i.e., ${\theta \in \mathbb{R}^n \colon \left\Vert \theta_0 -\theta \right\Vert_2\leq r}$. Clients whose models are parameterized by $\theta \in \mathbb{R}^n$ within the same $r$-neighbourhood are considered to be part of the same \emph{group} or \emph{cluster}.
\end{definition}
Algorithm~\ref{alg:pifca} is inspired by the Iterative Federated Clustering Algorithm (IFCA) proposed in~\cite{ghosh} and extends it by incorporating formal privacy guarantees. The key modifications include the introduction of the \texttt{SanitizeUpdate} function, as described in Algorithm~\ref{alg:sanitize}, and the utilization of $k$-means for server-side clustering of the updated models.

\begin{algorithm*}[htbp]
  \begin{algorithmic}
  \caption{An algorithm for personalized federated learning with formal privacy guarantees in local neighbourhoods.}
  \label{alg:pifca}
  \State Input: number of clusters $k$; initial hypotheses $\theta_j^{(0)},
  j \in \left\{1,\ldots, k \right\}$; number of rounds $T$; number of users per
  round $U$; number of local epochs $E$; local step size $s$;
  user batch size $B_s$; noise multiplier $\nu$; local dataset $Z_c$ held by user $c$.
  \For{$t = \left\{ 0, 1, \dots, T-1 \right\}$} \Comment{Server-side loop}
  \State $C^{(t)} \gets$ SampleUserSubset($U$)
  \State BroadcastParameterVectors($C^{(t)}$; $\theta_j^{(t)}, j \in \left\{1,\ldots, k \right\}$)
    \For{$c \in C^{(t)}$} in parallel \Comment{Client-side loop}
    \State $\bar{j} = \argmin_{j \in \left\{1,\ldots, k \right\}}F_c(\theta_j^{(t)};
    Z_c)$
    \State $\theta_{\bar{j}, c}^{(t)} \gets$ LocalUpdate($\theta_{\bar{j}}^{(t)};
      s; E; Z_c$)
      \State $\hat{\theta}_{\bar{j}, c}^{(t)} \gets$
      SanitizeUpdate($\theta_{\bar{j}, c}^{(t)}$; $\nu$)
    \EndFor
    \State $\left\{S_1, \dots, S_k\right\} = \text{k-means}$($\hat{\theta}_{\bar{j},c}^{(t)}$,
      $c \in C^{(t)}$; $\theta_j^{(t)}, j \in \left\{1,\ldots, k \right\}$)
    \State $\theta_j^{(t+1)} \gets \frac{1}{|S_j|}\sum_{c \in S_j}
    \hat{\theta}_{\bar{j}, c}^{(t)}, \quad \forall j \in \left\{1,\ldots, k \right\}$
  \EndFor
  \end{algorithmic}
\end{algorithm*}

\begin{algorithm}[htbp]
  \caption{SanitizeUpdate obfuscates a vector $\theta \in \mathbb{R}^n$,
  with a Laplacian noise tuned on the radius of a certain neighbourhood and centered
  in $0$.}
  \label{alg:sanitize}
  \begin{algorithmic}
    \Function{SanitizeUpdate}{$\theta_{\bar{j}}^{(t)}; \theta_{\bar{j}, c}^{(t)}; \nu$}
    \State $\delta_c^{(t)} = \theta_{\bar{j}, c}^{(t)} - \theta_{\bar{j}}^{(t)} $
    \State $\varepsilon = \frac{n}{\nu \Vert\delta_c^{(t)}\Vert}$
    \State Sample $\rho \sim \mathcal{L}_{0, \varepsilon}(x)$
    \State $\hat{\theta}_{\bar{j}, c}^{(t)}  = \theta_{\bar{j}, c}^{(t)} + \rho$
    \State \Return $\hat{\theta}_{\bar{j}, c}^{(t)}$
    \EndFunction
  \end{algorithmic}
\end{algorithm}

\subsection{The Laplace mechanism under Euclidean distance in
$\mathbb{R}^n$}~\label{sec:laplace}
The \texttt{SanitizeUpdate} function in Algorithm~\ref{alg:sanitize} is based on a generalization of the Laplace mechanism to $\mathbb{R}^n$ under the Euclidean distance, which was originally introduced in~\cite{geo} for geo-indistinguishability in $\mathbb{R}^2$. The decision to utilize the $L_2$ norm as the distance measure serves two main purposes.

First, clustering is performed on the vector space $\mathbb{R}^n$ of parameters, using the $k$-means algorithm, which relies on the Euclidean distance. By defining clusters or groups of users based on the proximity of their model parameters using the $L_2$ norm, the procedure needs a $d$-privacy mechanism that obscures the reported values within each group while enabling the server to distinguish among users belonging to different clusters.

Second, the use of equidistant noise vectors in the $L_2$ norm for sanitizing the parameters ensures equiprobability by construction. This property leads to the same bound on the increase of the cost function in first-order approximation, as demonstrated in Proposition~\ref{th:bound}. The Laplace mechanism under Euclidean distance in the general space $\mathbb{R}^n$ is formally defined in Proposition~\ref{lem:laplace}.

\begin{restatable}[]{proposition}{laplace}
\label{lem:laplace}
Let $\mathcal{L}_{\varepsilon}\colon\mathbb{R}^n \mapsto\mathbb{R}^n$ be the Laplace mechanism with distribution
  $\mathcal{L}_{x_0,\varepsilon}(x) = \mathbb{P}\left[\mathcal{L}_{\varepsilon}(x_0)=x\right] = Ke^{-\varepsilon d(x,x_0)}$ with $d(.)$ being the Euclidean distance. If $\rho \sim \mathcal{L}_{x_0, \varepsilon}(x)$, then:
  \begin{enumerate}
      \item   $\mathcal{L}_{x_0,\varepsilon}$ is $\varepsilon$-$d$-private and $K
                =\frac{\varepsilon^n\Gamma(\frac{n}{2})}{2\pi^{\frac{n}{2}}\Gamma(n)}$
      \item   $\norm{\rho}_2 \sim \gamma_{\varepsilon,n}(r) = \frac{\varepsilon^n e^{-\varepsilon r}
              r^{n-1}}{\Gamma(n)}$ \label{point:radius}
      \item   The $i^{th}$ component of $\rho$ has variance $\sigma_{\rho_i}^2 = \frac{n+1}{\varepsilon^2}$
  \end{enumerate}
\end{restatable}
where $\Gamma(n)$ is the Gamma function defined for positive reals as $\int_{0}^{\infty}t^{n-1}e^{-t}\, dt$ which reduces to the factorial function whenever $n \in \mathbb{N}$.
\begin{proof}
    The proof can be found in Appendix A of~\cite{Galli_Personalised_icissp23}.
\end{proof}

\begin{restatable}[]{proposition}{bound}
\label{th:bound}
Let $y = f(x,\theta)$ be the fitting function of a machine learning model parameterized by $\theta$, and $(X,Y) = Z$ the dataset over which the RMSE loss function $F(Z,\theta)$ is to be minimized, with $x\in X$ and $y \in Y$. If $\rho \sim \mathcal{L}_{0, \varepsilon}$, the bound on the increase of the cost function does not depend on the direction of $\rho$, in first-order approximation, and:
\begin{equation}
\begin{split}
    &\norm{F(Z,\theta + \rho)}_2 - \norm{F(Z,\theta)}_2 \leq 
 \\&\norm{J_f(X,\theta)}_2\norm{\rho}_2 + o(\norm{J_f(X,\theta)\cdot\rho}_2)
\end{split}
\end{equation}
\end{restatable}
\begin{proof}
    The proof can be found in Appendix A of~\cite{Galli_Personalised_icissp23}.
\end{proof}

The results in Proposition \ref{lem:laplace} allow to reduce the problem of sampling a point from Laplace to i) sampling the norm of such point according to the result in  Item \ref{point:radius} of Proposition \ref{lem:laplace} and then ii) sample uniformly a unit (directional) vector from the hypersphere in $\mathbb{R}^n$.
Much like DP, $d$-privacy provides a means to compute the total privacy parameters in case of repeated queries, a result known as the Compositionality Theorem for $d$-privacy.

\begin{restatable}[]{theorem}{compositionality}
\label{th:compositionality}
Let $\mathcal{K}_i$ be $(\varepsilon_i)$-$d$-private mechanism for $i\in \{1,2\}$. Then their independent composition is $(\varepsilon_1+\varepsilon_2)$-$d$-private.
\end{restatable}
\begin{proof}
    The proof can be found in Appendix A of~\cite{Galli_Personalised_icissp23}.
\end{proof}

\subsection{A Heuristic for defining the Neighbourhood of a Client} \label{heuristic}
During the $t$-th iteration, when a user $c$ invokes the \texttt{SanitizeUpdate} procedure in Algorithm~\ref{alg:sanitize},
it has already received a set of hypotheses, optimized $\theta_{\bar{j}}^{(t)}$ (the one that fits best its data distribution),
and got $\theta_{\bar{j}, c}^{(t)}$. It is reasonable to assume that clients whose 
datasets are sampled from the same underlying data distribution $\mathcal{D}_{\bar{j}}$ will perform an update similar to $\delta_c^{(t)}$.
Therefore, points which are within the $\delta_c^{(t)}$-neighbourhood of $\hat{\theta}_{\bar{j}, c}^{(t)}$ are forced to be indistinguishable. To provide this guarantee, the Laplace mechanism is tuned such that the points within the neighbourhood are $\varepsilon \Vert \delta_c^{(t)}\Vert_2$ differentially private.
By choosing $\varepsilon = n/(\nu \delta_c^{(t)})$, it results in  $\varepsilon \Vert \delta_c^{(t)}\Vert_2 = n/\nu$, where $\nu$ is referred to as the \emph{noise multiplier}. 
Notably, a larger value of $\nu$ corresponds to a stronger privacy guarantee. This is because the norm of the noise vector sampled from the Laplace distribution follows the distribution specified in Proposition \ref{lem:laplace}, with an expected value of $\mathbb{E}\left[ \gamma_{\varepsilon, n}(r)\right] = n/\varepsilon$.

\section{Experiments} \label{section:Experiments}
The following Section discusses a number of experimental validations of Algorithm \ref{alg:pifca} on different tasks and datasets. Detailed experimental settings are discussed in Appendix B of~\cite{Galli_Personalised_icissp23}, but we provide here an overview of the hardware and software stacks:
All the following experiments are run on a local server running Ubuntu 20.04.3 LTS with an AMD EPYC 7282 16-Core processor, 1.5TB of RAM and $8\times$ NVIDIA A100 GPUs. Python and PyTorch are the main software tools adopted for simulating the federation of clients and their corresponding collaborative training.

\subsection{Characterizing privacy} \label{section:ExperimentsPriv}
 In this Section, we aim to evaluate and assess the trade-off in training personalized federated learning models under formal local privacy guarantees.
\subsubsection{Synthetic Data}\label{sec:synthetic}
Data is generated according to $k=2$ different distributions: $y = x^T\theta_i^* + u$ and $u \sim \text{Uniform}\left[0, 1\right)$, $\forall i\in \{1,2\}$ and $\theta_1^* = \left[+5, +6\right]^T$, $\theta_2^* = \left[+4, -4.5\right]^T$.
We then assess how training progresses as we move from the Federated Averaging~\cite{fed-l-2} (Figure \ref{fig1:11}, \ref{fig1:12}, \ref{fig1:13}), to IFCA (Figure \ref{fig1:21}, \ref{fig1:22}, \ref{fig1:23}), and finally Algorithm \ref{alg:pifca} (Figure \ref{fig1:31}, \ref{fig1:32}, \ref{fig1:33}). When utilizing Federated Averaging, a noticeable issue arises: relying on a single hypothesis fails to capture the diversity present in the data distributions. As a result, the final parameters tend to settle somewhere between the optimal parameter values (see Figure \ref{fig1:12}). Conversely, employing IFCA demonstrates that having multiple initial hypotheses enhances performance, particularly when clients possess heterogeneous data. This is evident from the nearly overlapping optimized client parameters with the true optimal parameters (see Figure \ref{fig1:22}).

By adopting our algorithm instead, not only do we provide formal guarantees, but we also achieve remarkable outcomes in terms of proximity to the optimal parameters (see Figure \ref{fig1:32}) and reduction of the loss function (see Figure \ref{fig1:33}). To assess privacy infringement, Figure \ref{priv-leak-synth} illustrates the maximum level of privacy leakage incurred by clients per cluster.

\newcommand{\mywidth}{0.33}
\begin{figure*}[htbp]
 \centering
 \captionsetup[subfigure]{justification=centering}
 \begin{subfigure}[b]{\mywidth\textwidth}
     \centering
     \includegraphics[width=\textwidth]{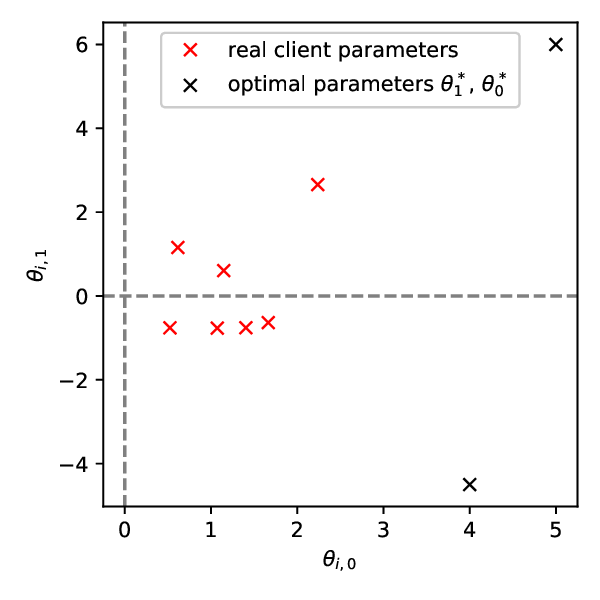}
     \caption{First round}
     \label{fig1:11}
 \end{subfigure}
 \hspace{-0.8em} 
 \begin{subfigure}[b]{\mywidth\textwidth}
     \centering
     \includegraphics[width=\textwidth]{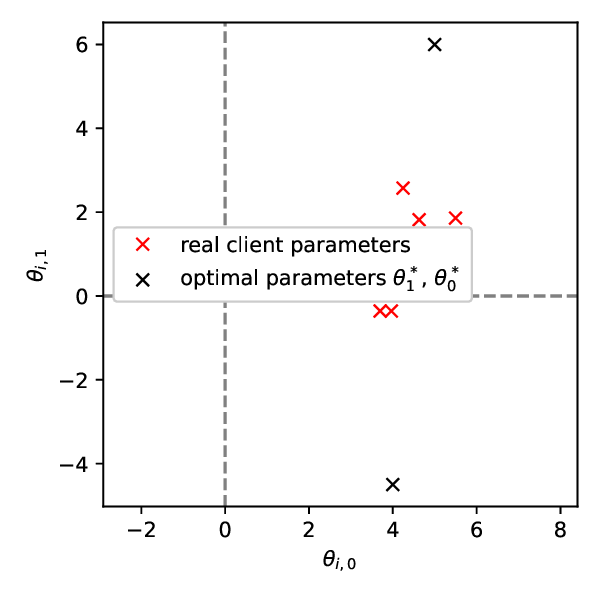}
     \caption{Best round}
     \label{fig1:12}
 \end{subfigure}
 \hspace{-0.8em} 
 \begin{subfigure}[b]{\mywidth\textwidth}
     \centering
     \includegraphics[width=\textwidth]{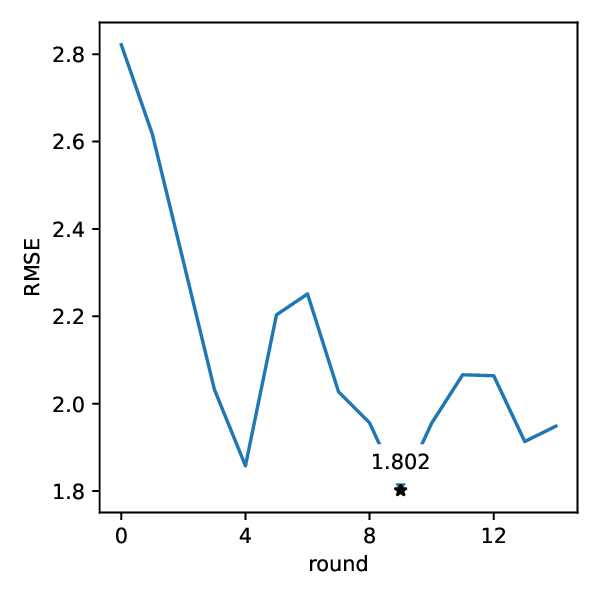}
     \caption{Validation loss}
     \label{fig1:13}
 \end{subfigure}

\medskip
 \begin{subfigure}[b]{\mywidth\textwidth}
     \centering
     \includegraphics[width=\textwidth]{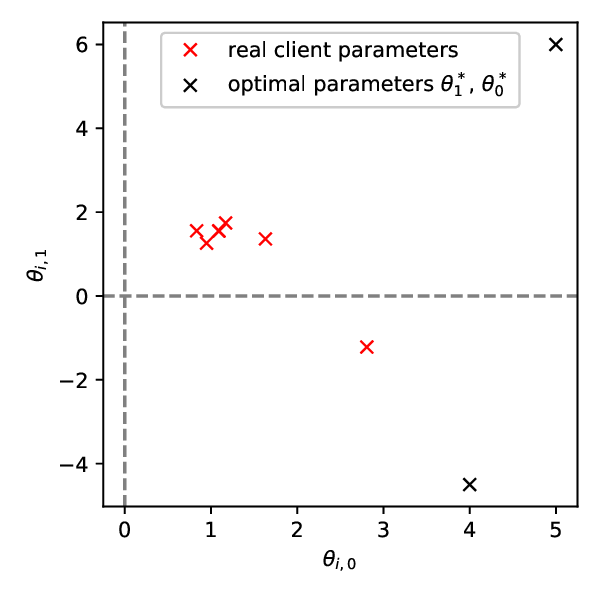}
     \caption{First round}
     \label{fig1:21}
 \end{subfigure}
 \hspace{-0.8em}
 \begin{subfigure}[b]{\mywidth\textwidth}
     \centering
     \includegraphics[width=\textwidth]{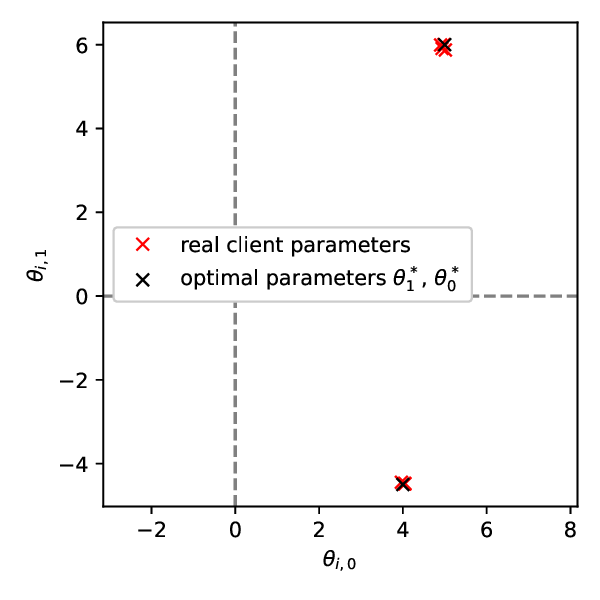}
     \caption{Best round}
     \label{fig1:22}
 \end{subfigure}
 \hspace{-0.8em}
 \begin{subfigure}[b]{\mywidth\textwidth}
     \centering
     \includegraphics[width=\textwidth]{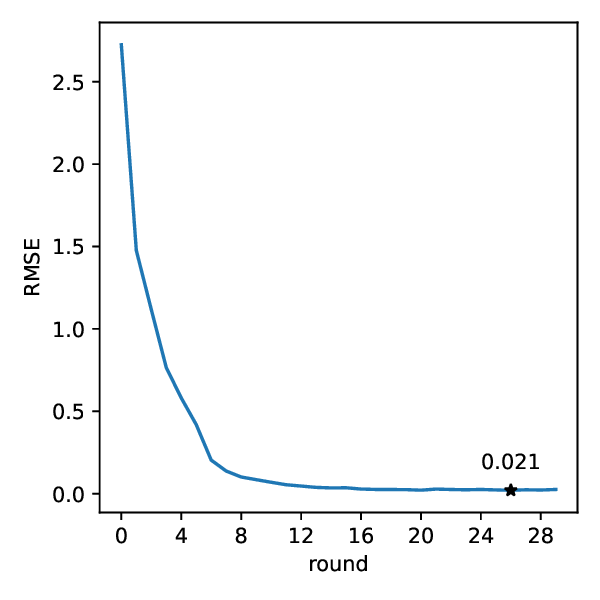}
     \caption{Validation loss}
     \label{fig1:23}
 \end{subfigure}
\medskip

 \begin{subfigure}[b]{\mywidth\textwidth}
     \centering
     \includegraphics[width=\textwidth]{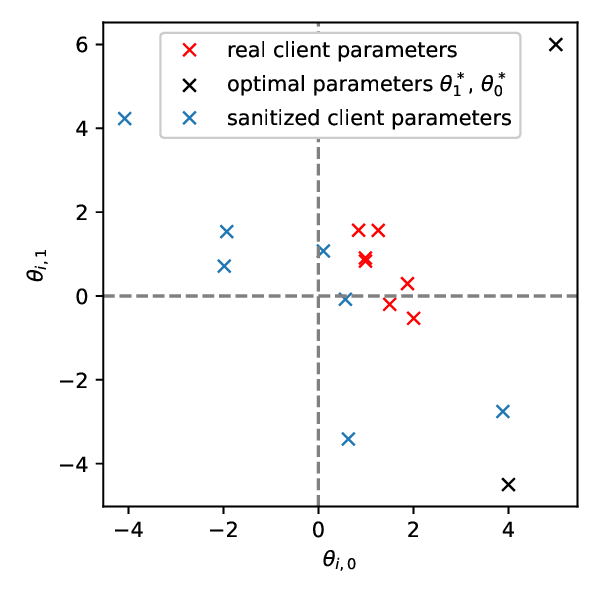}
     \caption{First round}
     \label{fig1:31}
 \end{subfigure}
 \hspace{-0.8em}
 \begin{subfigure}[b]{\mywidth\textwidth}
     \centering
     \includegraphics[width=\textwidth]{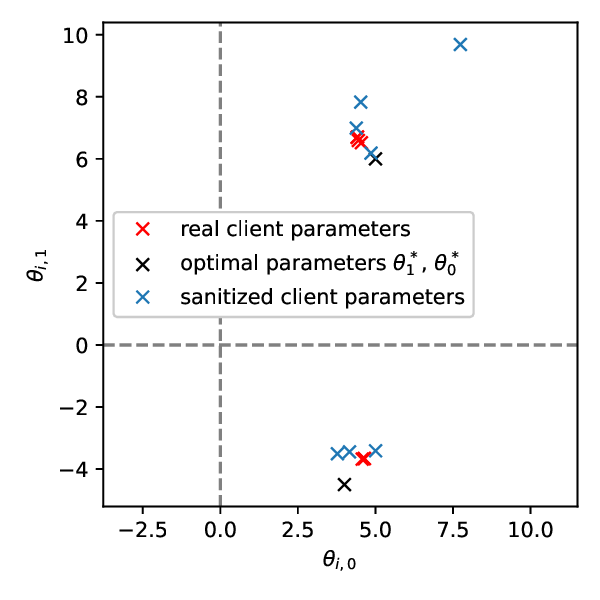}
     \caption{Best round}
     \label{fig1:32}
 \end{subfigure}
 \hspace{-0.8em}
 \begin{subfigure}[b]{\mywidth\textwidth}
     \centering
     \includegraphics[width=\textwidth]{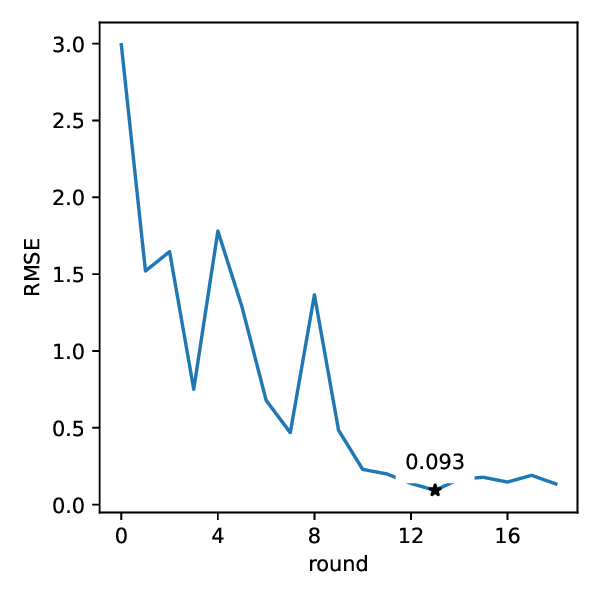}
     \caption{Validation loss}
     \label{fig1:33}
 \end{subfigure}
 \medskip
 \caption{(From~\cite{Galli_Personalised_icissp23}) Learning federated linear models with: (a, b, c) one initial hypothesis and
   non-sanitized communication, (d, e, f) two initial hypotheses and non-sanitized communication,
   (g, h, i) two initial hypotheses and sanitized communication. The first
 two figures of each row show the parameter vectors released by the clients
 to the server.}
    \label{fig1}
\end{figure*}

\begin{figure}[htbp]
    \centering
    \includegraphics[scale=0.48]{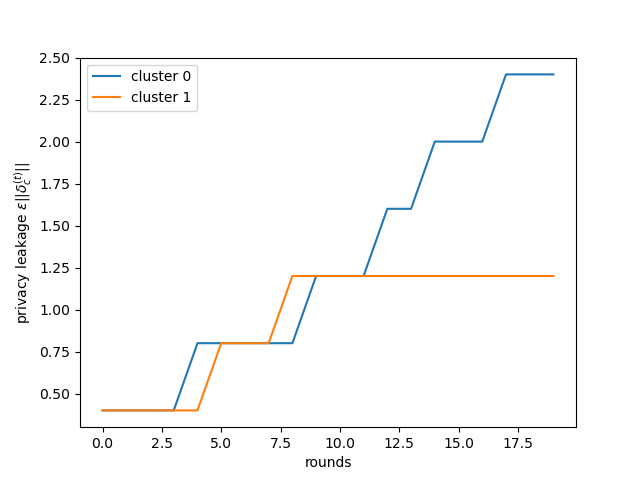}
    \caption{(From~\cite{Galli_Personalised_icissp23}) Synthetic data: max privacy leakage among clients. Privacy leakage is constant when clients with the largest privacy leakage are not sampled (by chance) to participate in those rounds.}
    \label{priv-leak-synth}
\end{figure}

\subsubsection{Hospital Charge Data}\label{sec:hospital}
This experiment utilizes the Hospital Charge Dataset obtained from the Centers for Medicare and Medicaid Services of the US Government~\cite{CMMS}. Here, the healthcare providers are regarded as the clients who participate in training a machine learning model through federated learning. The objective is to predict the cost of a medical service based on its location in the country and the specific procedure involved.

\begin{figure}[htbp]
    \centering
    \includegraphics[scale=0.53]{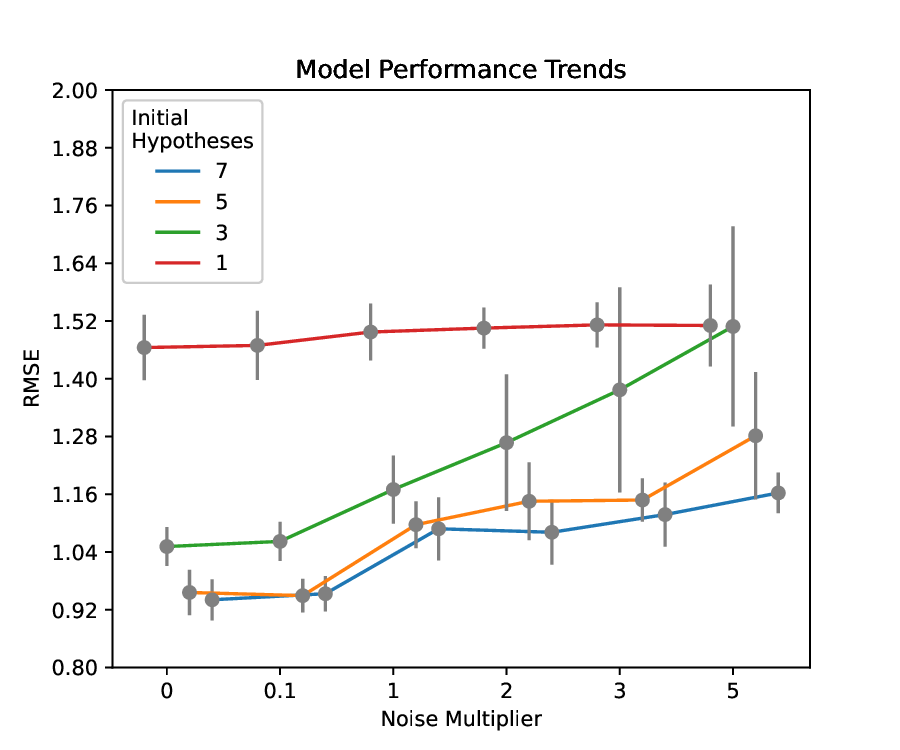}
    \caption{(From~\cite{Galli_Personalised_icissp23}) RMSE for models trained with Algorithm \ref{alg:pifca} on the Hospital 
    Charge Dataset. Error bars show $\pm \sigma$, with $\sigma$ the empirical standard
    deviation. Lower RMSE values are better for accuracy.}
    \label{fig:hospital}
\end{figure}

\begin{figure}[htbp]
        \centering
        \includegraphics[scale=0.48]{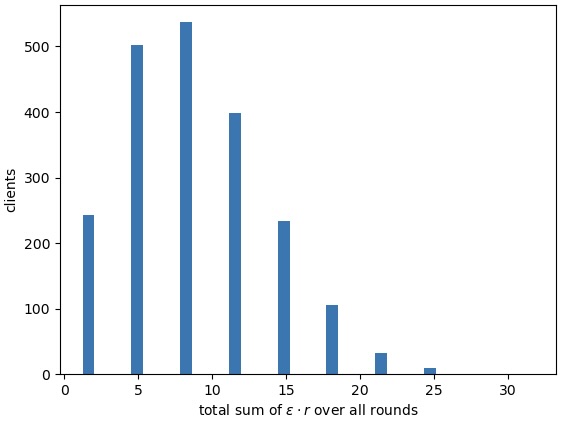}
        \caption{(From~\cite{Galli_Personalised_icissp23}) Hospital charge data: the empirical distribution of the privacy budget over the clients for $\nu=3$, 5 initial hypotheses, seed $=3$, $r$ is the radius of the neighbourhood,  the total number of clients is 2062.}
    \label{priv-leak-hospital}
\end{figure}

To evaluate the trade-off between privacy, personalization, and accuracy, we explore various numbers of initial hypotheses since the number of underlying data distributions is unknown a priori. Accuracy is assessed at different levels of the noise multiplier $\nu$. Notably, using Algorithm \ref{alg:pifca} with only one hypothesis yields the Federated Averaging algorithm. As depicted in Figure \ref{fig:hospital}, employing multiple hypotheses significantly reduces the RMSE loss function, particularly when transitioning from one to three hypotheses. Furthermore, we emphasize that increasing the number of hypotheses also helps mitigate the impact of the noise multiplier, even at high levels (as shown on the right-hand side of the figure). This highlights the importance of adopting formal privacy guarantees when a slight increase in the cost function is acceptable. The empirical distribution of privacy leakage among clients involved in a specific training configuration is illustrated in Figure \ref{priv-leak-hospital}. Table \ref{table:privacy-budget-correct} presents privacy leakage statistics across multiple rounds and configurations.

\begin{table*}[t]
    \begin{center}
        \begin{tabular}{*5c}
            \toprule
            & \multicolumn{4}{c}{Hypotheses}  \\
            \midrule
            $\nu$ & $7$ & $5$ & $3$ & $1$ \\
            \midrule
            0 & $-, -$ & $-, -$ & $-, -$ & $-, -$ \\
            0.1 & 517.0, 1551.0 & 418.0, 1342.0 & 473.0, 1386.0 & 528.0, 1540.0 \\
            1 & 36.3, 126.5 & 40.7, 127.6 & 44.0, 138.6 & 49.5, 147.4 \\
            2 & 15.4, 57.8 & 14.3, 54.5 & 22.0, 69.3 & 21.5, 66.6 \\
            3 & 7.7, 32.3 & 8.4, 36.7 & 12.5, 40.0 & 12.1, 40.0 \\
            5 & 5.7, 21.3 & 5.9, 22.0 & 5.5, 21.6 & 5.3, 20.9 \\
            \bottomrule
        \end{tabular}
    \end{center}
    \caption{(From~\cite{Galli_Personalised_icissp23}) Hospital charge data: median and maximum local privacy budgets over the whole set of clients, averaged over $10$ runs with different seeds. $\nu=0$ means no privacy guarantee.}
    \label{table:privacy-budget-correct}
\end{table*}

\subsubsection{FEMNIST Image Classification}~\label{nn}

\begin{table*}[htbp]
\begin{center}
\begin{tabular}{ccccc}
\toprule
& \multicolumn{2}{c}{Cross Entropy loss}
& \multicolumn{2}{c}{RMSE loss}\\ \hline
\begin{tabular}[c]{@{}c@{}} $\nu$\\ \end{tabular}  & \begin{tabular}[c]{@{}c@{}}Average \\ Accuracy\end{tabular} & \begin{tabular}[c]{@{}c@{}}Standard\\ Deviation\end{tabular} & \begin{tabular}[c]{@{}c@{}}Average\\ Accuracy\end{tabular} & \begin{tabular}[c]{@{}c@{}}Standard\\ Deviation\end{tabular} \\ \hline
0                & 0.832                                                       & $\pm$ 0.012                                                        & 0.801                                                      & $\pm$ 0.001                                                        \\
0.001            & 0.843                                                       & $\pm$ 0.006                                                        & 0.813                                                      & $\pm$ 0.014                                                        \\
0.01             & 0.832                                                       & $\pm$ 0.017                                                        & 0.805                                                      & $\pm$ 0.008                                                        \\
0.1              & 0.834                                                       & $\pm$ 0.026                                                        & 0.808                                                      & $\pm$ 0.019                                                        \\
1                & 0.834                                                       & $\pm$ 0.014                                                        & 0.814                                                      & $\pm$ 0.012                                                        \\
3                & 0.835                                                       & $\pm$ 0.017                                                        & 0.825                                                      & $\pm$ 0.010                                                        \\
5                & 0.812                                                       & $\pm$ 0.016                                                        & 0.787                                                      & $\pm$ 0.003                                                        \\
10               & 0.692                                                       & $\pm$ 0.002                                                        & 0.687                                                      & $\pm$ 0.014                                                        \\
15               & 0.561                                                       & $\pm$ 0.005                                                        & 0.622                                                      & $\pm$ 0.003                                                        \\ 
\bottomrule
\end{tabular}
\end{center}
\caption{(From~\cite{Galli_Personalised_icissp23}) Effects of increasing the noise multiplier on the
validation accuracy and standard deviation.} \label{nn-results}
\end{table*}

\begin{figure}[htbp]
    \centering
    \includegraphics[scale=0.62]{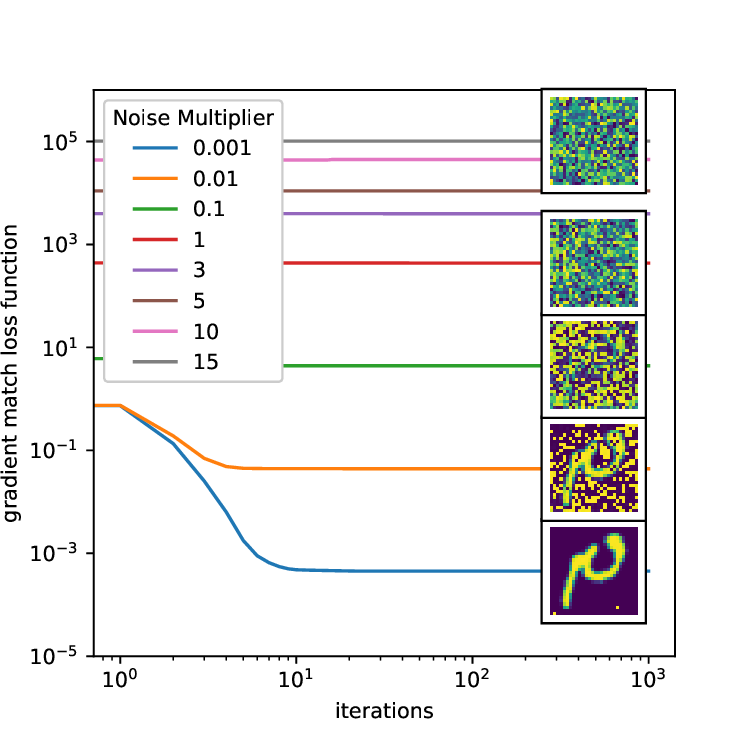}
    \caption{(From~\cite{Galli_Personalised_icissp23}) Effects of the Laplace mechanism in Proposition \ref{lem:laplace} with different noise multipliers as a defence strategy against the DLG attack.}\label{fig:dlg}
\end{figure}

This task involves character recognition from images using the FEMNIST dataset~\cite{caldas2018leaf}.
When selecting the range of noise multipliers $\nu$, the resulting privacy leakage $\varepsilon \Vert \delta_c^{(t)} \Vert_2 = n/\nu$ would be exceptionally large, given the CNN's $n=206590$ parameters. Consequently, this renders the mechanism incapable of providing meaningful theoretical privacy guarantees. This issue is commonly encountered with local privacy mechanisms~\cite{ldp-bounds}, as the expected value of the noise vector's norm, $\mathbb{E}\left[ \gamma_{\varepsilon, n}(r)\right]$, exhibits a linear dependence on $n$: $n/\varepsilon$.

However, it is still possible to evaluate, in practice, whether this specific generalization of the Laplace mechanism can effectively defend against a particular attack known as DLG~\cite{deep-leakage}. The outcomes of varying noise multiplier values are presented in Figure \ref{fig:dlg}, and Table \ref{nn-results} provides additional details. Notably, when $\nu = 10^{-3}$, the ground truth image can be fully reconstructed. Partial reconstruction remains possible up to $\nu = 10^{-1}$. However, for $\nu \geq 1$, experimental results demonstrate the failure of the DLG attack to reconstruct input samples when the communication between the client and server is protected by the mechanism outlined in Proposition \ref{lem:laplace}.

\subsection{Fairness analysis}~\label{section:ExperimentsFairness}
In this section, we analyze how group fairness improves with the personalization of the trained models under $d$-privacy guarantees when there are two groups with different data distributions. Experiments were performed on synthetic data and the FEMNIST image classification dataset that was used in Section \ref{section:ExperimentsPriv}. To ensure a thorough evaluation, we considered a variety of group fairness metrics in the experiments. In particular, we measured the fairness with respect to equal opportunity~\cite{hardt2016equality}, equalized odds~\cite{hardt2016equality}, and demographic parity~\cite{dwork2012fairness} as explained in Section~\ref{section:BackgroundFairness}.

In particular, in Figures~\ref{fig:synthetic_fair} and \ref{fig:femnist_fair}, the $X$-axis denotes the noise multiplier $\nu$ representing the amount of $d$-private noise added to the local updates as explained in Section \ref{heuristic} and the $Y$-axis denotes the absolute value of the difference in fairness between the privileged and unprivileged groups with respect to the different metrics of group fairness that we considered. 

\subsubsection{Synthetic data}\label{section:ExperimentsFairness_synthetic}

\newcommand{\fewidth}{0.5}
\begin{figure*}[htbp]
 \centering
 \captionsetup[subfigure]{justification=centering}
 \begin{subfigure}[b]{\fewidth\textwidth}
     \centering
     \includegraphics[width=\textwidth]{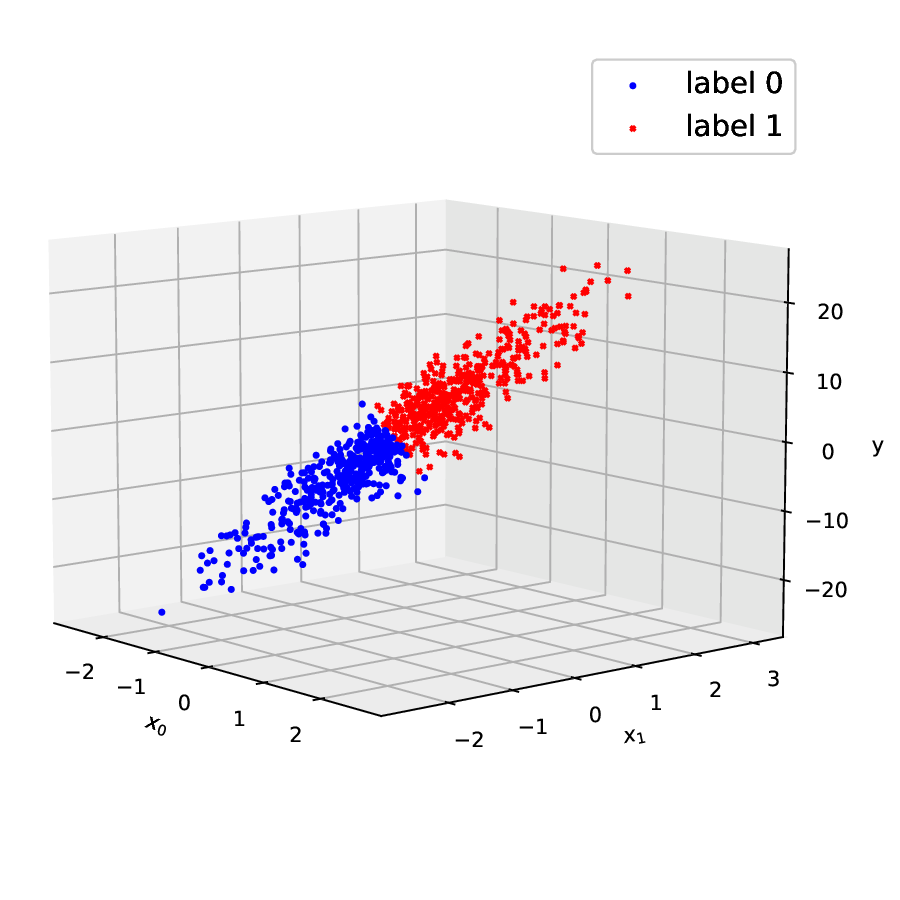}
     \label{fig6:1}
 \end{subfigure}
 \hspace{-1em}  
 \begin{subfigure}[b]{\fewidth\textwidth}
     \centering
     \includegraphics[width=\textwidth]{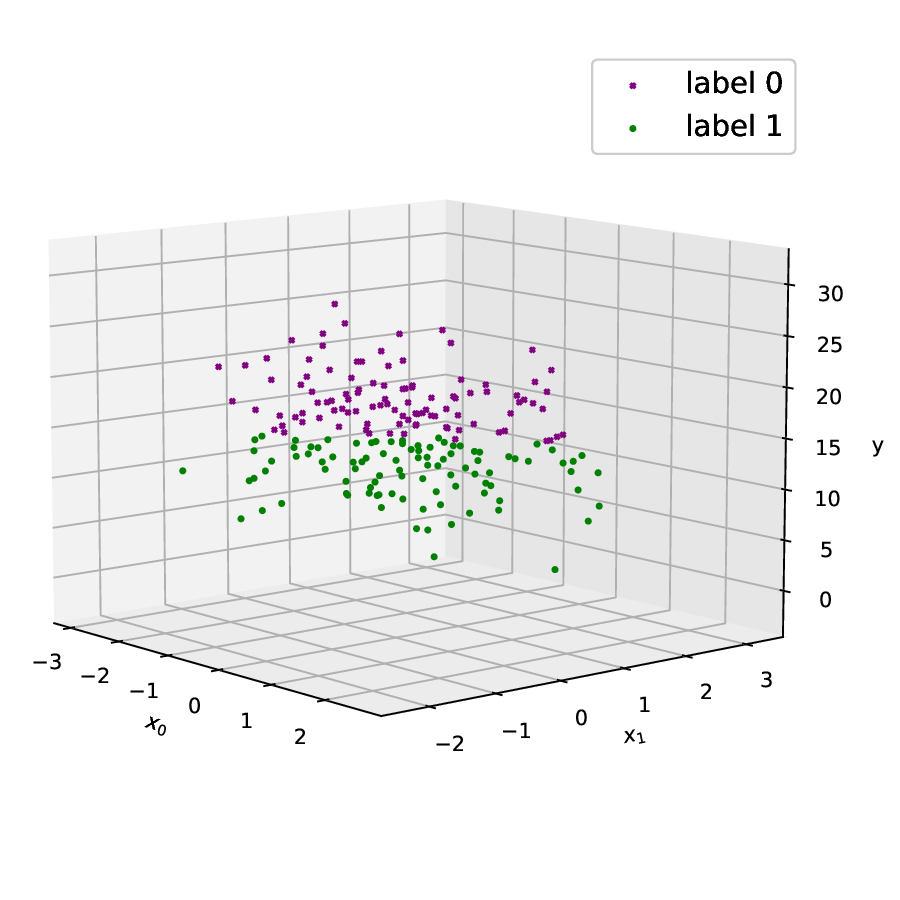}
     \label{fig6:2}
 \end{subfigure}
 \begin{subfigure}[b]{\fewidth\textwidth}
     \centering
     \includegraphics[width=\textwidth]{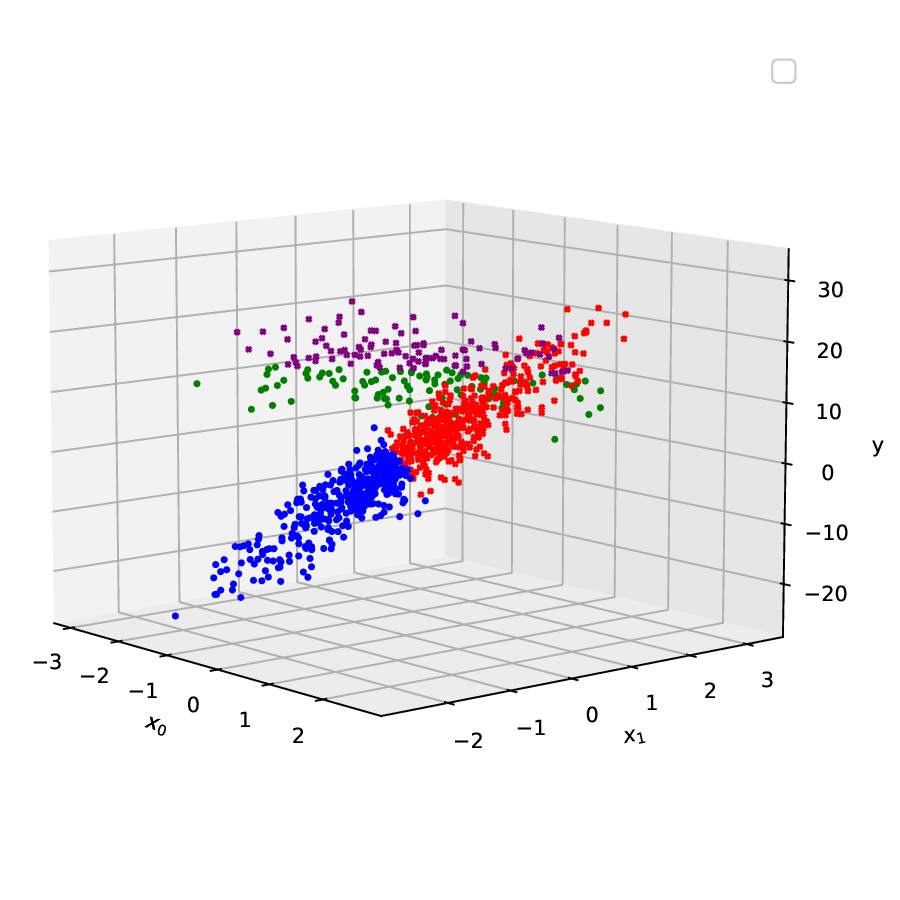}
     \label{fig6:3}
 \end{subfigure}
 \hspace{-1em} 

\caption{The first two plots from the left illustrate the spatial distribution of the samples in $g_1$ and $g_2$, respectively, and the third plot shows $g_1$ and $g_2$ superimposed together in the same space.}~\label{fig:all}
\end{figure*}

\color{black}
Synthetic data was generated in a method similar to that in Section~\ref{sec:synthetic} with the following modifications to enable us to investigate the aspect of group fairness fostered by our method: 
i) Total number of users is $1000$ and each user holds $10$ samples. $800$ users have data that is generated according to distributions $y = x^T\theta_1 + u$ and $u \sim \text{Uniform}\left[0, 1\right)$, $\forall i\in \{1,2\}$, and set as a privileged majority group $g_1$. The remaining 200 users have data that is generated according to distribution $y = x^T\theta_2 +15 + u$ and $u \sim \text{Uniform}\left[0, 1\right)$, $\forall i\in \{1,2\}$, and set as an unprivileged minority group $g_2$. In this case, the sensitive attribute considered to evaluate fairness is the group id $G$ where $G\in \{g_1, g_2\}$.
ii) For binary classification, we set labels by using the $z=\operatorname{Sigmoid}(Y), \,\forall\,  y, \hat{y} \in Y$. In the case of $g_1$, we assign the label $1$ if the value of $z$ is greater than or equal to $0.5$ and assign the label $0$ otherwise. On the other hand, in the case of $g_2$, the label $1$ is assigned when the $z=\operatorname{Sigmoid}(Y-15), \,\forall\,  y, \hat{y} \in Y$ is less than or equal to $0.5$, and the label $0$ is assigned otherwise. This setting is to simulate a situation in which discrimination occurs depending on sensitive attributes in the real world such as minorities would have experienced a higher loan rejection rate than white applicants with the same property~\cite{bartlett2022consumer}. Thus, in our experiment, label $1$ could be interpreted as ``loan approved'' and label $0$ as ``loan denied''. The data generated in this way are shown in Figure~\ref{fig:all}.

We compared the fairness for two cases: one with a single hypothesis (no personalization) and the other with the number of hypotheses as 2 (with personalization) in the framework of Algorithm~\ref{alg:pifca}. The experimental results are demonstrated in Figure~\ref{fig:synthetic_fair}.

\newcommand{\sywidth}{0.5}
\begin{figure*}[htbp]
 \centering
 \captionsetup[subfigure]{justification=centering}
 \begin{subfigure}[b]{\sywidth\textwidth}
     \centering
     \includegraphics[width=\textwidth]{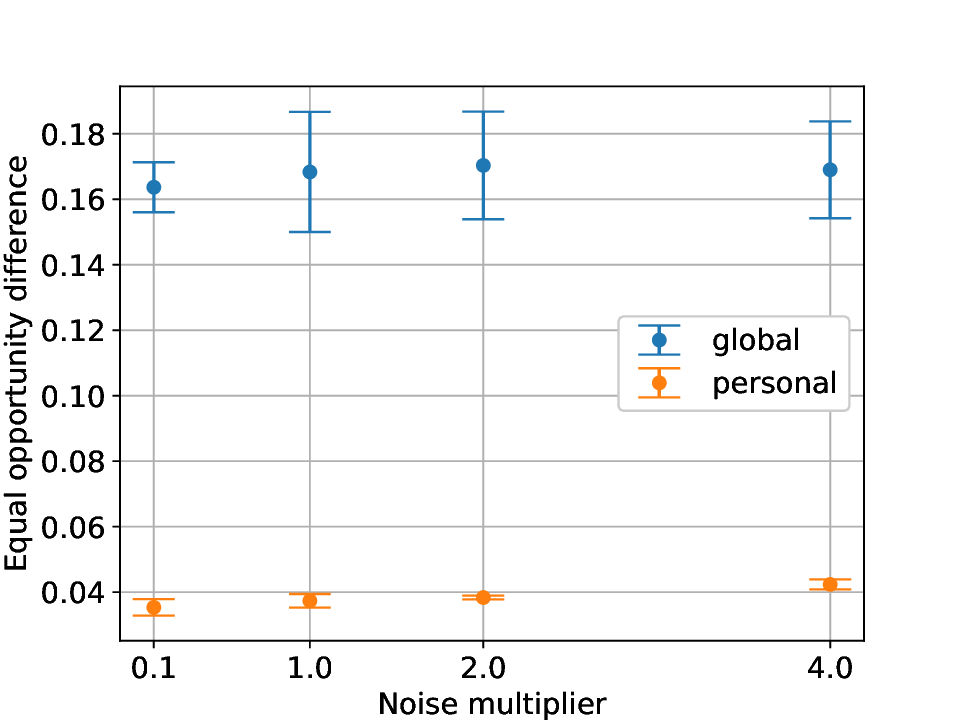}
     \label{fig:synthetic_EqOpp}
          \caption{Equal opportunity difference}
 \end{subfigure}
 \hspace{-1em} 
 \begin{subfigure}[b]{\sywidth\textwidth}
     \centering
     \includegraphics[width=\textwidth]{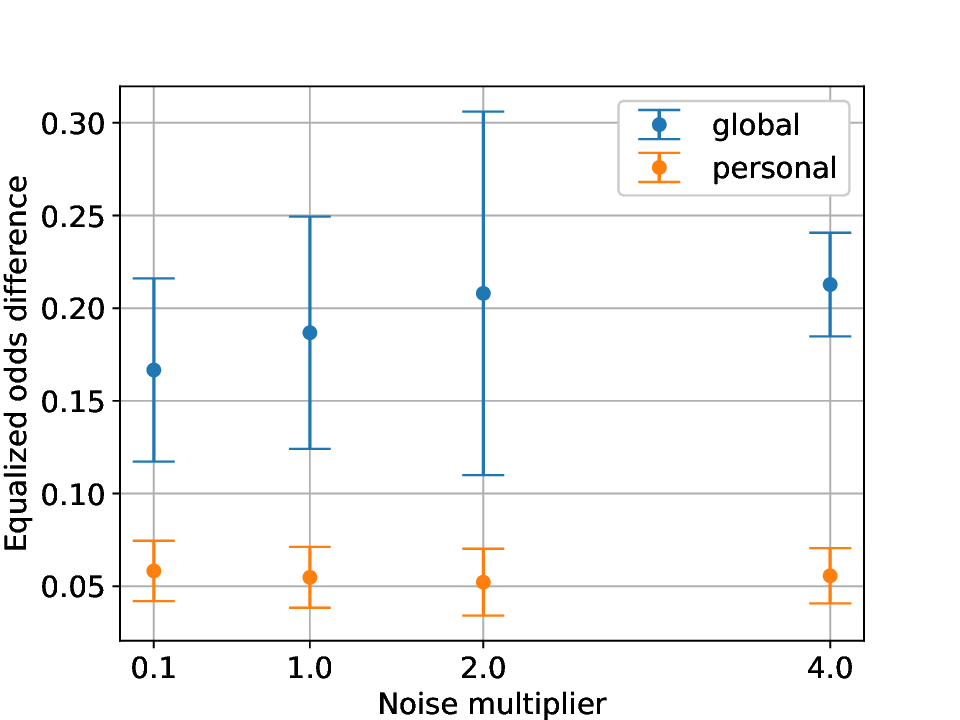}
     \label{fig:synthetic_EqOdds}
     \caption{Equalized odds difference}
 \end{subfigure}
 \hspace{-1em} 
 \begin{subfigure}[b]{\sywidth\textwidth}
     \centering
     \includegraphics[width=\textwidth]{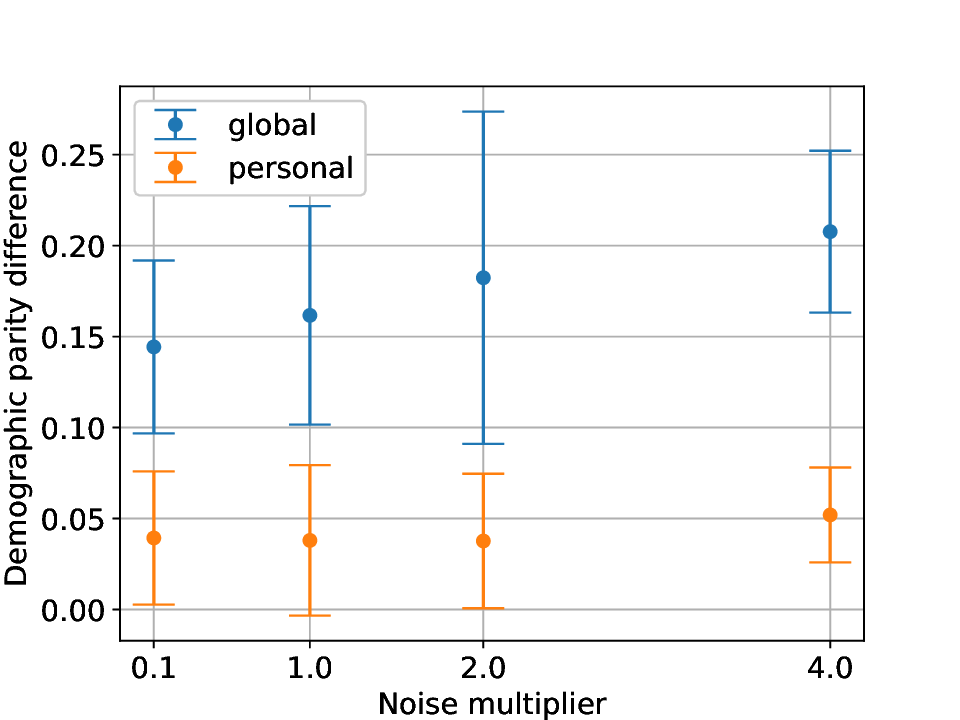}
     \label{fig:synthetic_DemParity}
     \caption{Demographic parity difference}
 \end{subfigure}
 \caption{The figure shows the comparison between the personalized and non-personalized models for (from left) equal opportunity, equalized odds, and demographic parity, respectively. Experiments were performed for noise multipliers $\nu$ of 0.1, 1, 2, and 4. For all the metrics of fairness and the values of the noise multiplier, the personalized model is seen to possess improved fairness over the non-personalized model.}
    ~\label{fig:synthetic_fair}
\end{figure*}

The results illustrated by Figure \ref{fig:synthetic_fair} assert that the personalization of models (i.e., Algorithm~\ref{alg:pifca}) enhances the group fairness under all the metrics and the levels of formal privacy guarantees compared to that of the non-personalized model. A major reason behind this significant improvement of fairness by the personalized model is that unlike the non-personalized model, which trains using data from both groups that are biased towards the majority group $g_1$, the personalized model training optimizes for each group’s data distribution without disregarding the effect of the minority group $g_2$. 
We also observe that fairness deteriorates as the value of the noise multiplier increases, as we would expect. This is presumably due to the decreasing influence of the minority group $g_2$ as the amount of noise insertion increases. This is consistent with the philosophy behind and the definition of DP and its variants. Furthermore, interestingly we observe that the personalized model ensures better fairness than the non-personalized model even with the highest level of privacy protection. This shows that personalization in FL under $d$-privacy can be a comprehensive solution towards privacy-preserving and ethical machine learning as it provides both privacy guarantees and enhanced fairness.

\subsubsection{FEMNIST Image Classification}~\label{section:ExperimentsFairness_FEMNIST}

\newcommand{\fmwidth}{0.5}
\begin{figure*}[htbp]
 \centering
 \captionsetup[subfigure]{justification=centering}
 \begin{subfigure}[b]{\fmwidth\textwidth}
     \centering
     \includegraphics[width=\textwidth]{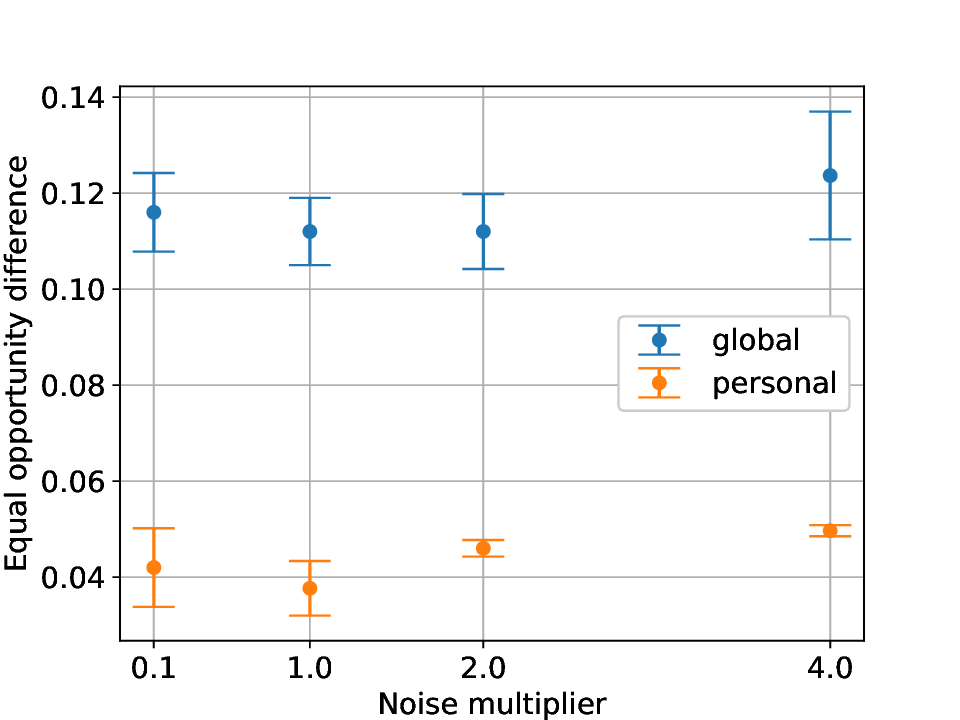}
     \label{fig:FEMNIST_EqOpp}
          \caption{Equal opportunity difference}
 \end{subfigure}
 \hspace{-1em} 
 \begin{subfigure}[b]{\fmwidth\textwidth}
     \centering
     \includegraphics[width=\textwidth]{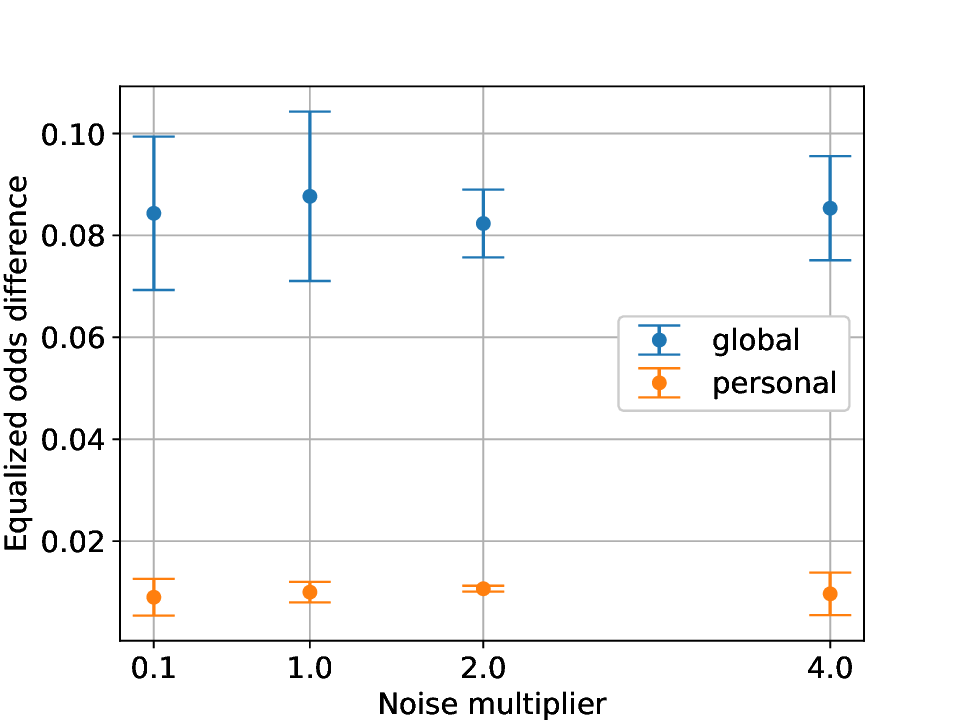}
     \label{fig:FEMNIST_EqOdds}
     \caption{Equalized odds difference}
 \end{subfigure}
 \hspace{-1em} 
 \begin{subfigure}[b]{\fmwidth\textwidth}
     \centering
     \includegraphics[width=\textwidth]{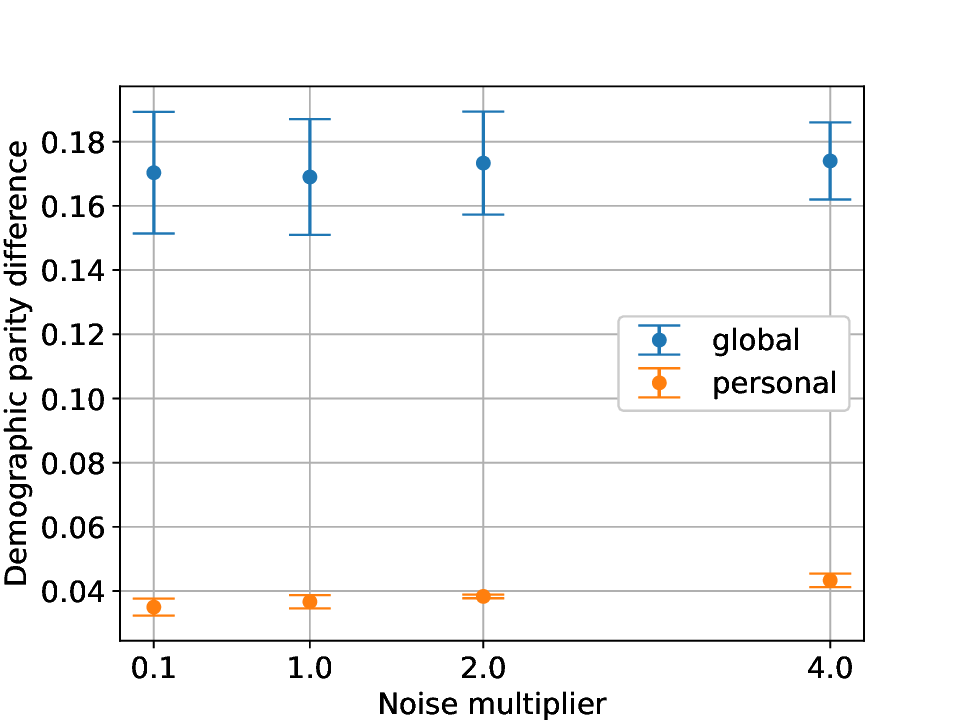}
     \label{fig:FEMNIST_DemParity}
     \caption{Demographic parity difference}
 \end{subfigure}
    \caption{The figure shows the comparison between the personalized and non-personalized models for equal opportunity equalized odds, and demographic parity. Experiments were performed for noise multipliers $\nu$ of 0.1, 1, 2, and 4. For all metrics of fairness and values of the noise multiplier, the personalized model improved fairness over the non-personalized model.}~\label{fig:femnist_fair}
\end{figure*}

To evaluate the fairness of our method on real datasets, we considered FEMNIST image classification dataset in the same form as in Section~\ref{nn}. As in experiments performed with the synthetic data in Section~\ref{section:ExperimentsFairness_synthetic}, the size of the groups considered privileged and unprivileged were different denoting the existence of a majority and a minority in the population. In this part, the rotated images are set as the unprivileged group $g_2$ with a total number of sampled users of $382$ forming only $20\%$ of all users. 
and the un-rotated images are used to represent the privileged group $g_1$ 
with a total number of users of $1736$. Like in the case of synthetic data considered before, the group membership was used to denote the sensitive attribute. In the case of $g_1$, we assign label $1$ if the FEMNIST image label is even and $0$ if it is odd. And for the $g_2$, we assign label $0$ if the FEMNIST image label is even and assign $1$ if it is odd. The experimental results are given by Figure~\ref{fig:femnist_fair}.

We observe that the personalized model training harbours significantly better group fairness across all metrics compared to its non-personalized counterpart. The change in fairness due to the amount of noise added was not as notable as in the case of the synthetic dataset but it was still observed to deteriorate with an increase in the value of the noise multiplier. Personalized model training in FL under the highest level of privacy is still observed to have better fairness across all the metrics than (non-personalized) models trained in a classical FL framework even with no privacy, similar to what we observed in the experiments with the synthetic data.

\section{Conclusion} \label{section:Conclusion}
This work builds upon our previous research on personalized federated learning with metric privacy guarantees. To ensure the privacy of ML model parameters during transmission, we employ $d$-privacy techniques for sanitization. The objective of this process is to generate personalized models that converge to optimal parameters, catering to the diverse datasets present in the federated learning setting.
Given the presence of multiple, unknown data distributions among the individuals participating in the federated learning process, we make a reasonable assumption of a mixture of these distributions. To effectively aggregate clients with similar data distributions, we employ a clustering approach using $k$-means on the sanitized parameter vectors. This method proves suitable because $d$-private mechanisms preserve the underlying topology of the true value domain. Notably, our mechanism shows particular promise for machine learning models with a relatively small number of parameters. Although the formal privacy guarantees diminish with larger models, experimental results demonstrate the effectiveness of the Laplace mechanism against the DLG attack.

In addition to metric privacy guarantees, we also evaluate the fairness of machine learning models trained using personalized federated learning and $d$-privacy. Our study assesses various group fairness metrics, including equal opportunity, equalized odds, and demographic parity. The consistent findings demonstrate that personalized models significantly improve group fairness across all evaluated metrics and privacy levels. Moreover, they, unlike non-personalized models, optimize for each group's specific data distribution, effectively mitigating biases towards the majority group. Consequently, significant advancements in fairness are achieved through this approach.

The level of fairness is influenced by the incorporation of $d$-private noise in the local updates. 
As the noise 
increases, the influence of the minority group decreases, resulting in a deterioration of fairness. This behaviour aligns with the principles of differential privacy and the expected impact of noise addition on group fairness. Remarkably, even with the highest level of privacy protection, personalized models consistently maintain superior fairness compared to non-personalized models. This observation highlights the potential of personalized model training in federated learning under $d$-privacy as a comprehensive solution for privacy-preserving and ethical machine learning. By offering privacy guarantees alongside enhanced fairness, personalized models demonstrate their effectiveness in balancing these critical aspects.

\section{Conflict of Interest}
On behalf of all authors, the corresponding author states that there is no conflict of interest.

\bibliography{sn-bibliography}


\begin{thebibliography}{55}
\ifx \bisbn   \undefined \def \bisbn  #1{ISBN #1}\fi
\ifx \binits  \undefined \def \binits#1{#1}\fi
\ifx \bauthor  \undefined \def \bauthor#1{#1}\fi
\ifx \batitle  \undefined \def \batitle#1{#1}\fi
\ifx \bjtitle  \undefined \def \bjtitle#1{#1}\fi
\ifx \bvolume  \undefined \def \bvolume#1{\textbf{#1}}\fi
\ifx \byear  \undefined \def \byear#1{#1}\fi
\ifx \bissue  \undefined \def \bissue#1{#1}\fi
\ifx \bfpage  \undefined \def \bfpage#1{#1}\fi
\ifx \blpage  \undefined \def \blpage #1{#1}\fi
\ifx \burl  \undefined \def \burl#1{\textsf{#1}}\fi
\ifx \doiurl  \undefined \def \doiurl#1{\url{https://doi.org/#1}}\fi
\ifx \betal  \undefined \def \betal{\textit{et al.}}\fi
\ifx \binstitute  \undefined \def \binstitute#1{#1}\fi
\ifx \binstitutionaled  \undefined \def \binstitutionaled#1{#1}\fi
\ifx \bctitle  \undefined \def \bctitle#1{#1}\fi
\ifx \beditor  \undefined \def \beditor#1{#1}\fi
\ifx \bpublisher  \undefined \def \bpublisher#1{#1}\fi
\ifx \bbtitle  \undefined \def \bbtitle#1{#1}\fi
\ifx \bedition  \undefined \def \bedition#1{#1}\fi
\ifx \bseriesno  \undefined \def \bseriesno#1{#1}\fi
\ifx \blocation  \undefined \def \blocation#1{#1}\fi
\ifx \bsertitle  \undefined \def \bsertitle#1{#1}\fi
\ifx \bsnm \undefined \def \bsnm#1{#1}\fi
\ifx \bsuffix \undefined \def \bsuffix#1{#1}\fi
\ifx \bparticle \undefined \def \bparticle#1{#1}\fi
\ifx \barticle \undefined \def \barticle#1{#1}\fi
\bibcommenthead
\ifx \bconfdate \undefined \def \bconfdate #1{#1}\fi
\ifx \botherref \undefined \def \botherref #1{#1}\fi
\ifx \url \undefined \def \url#1{\textsf{#1}}\fi
\ifx \bchapter \undefined \def \bchapter#1{#1}\fi
\ifx \bbook \undefined \def \bbook#1{#1}\fi
\ifx \bcomment \undefined \def \bcomment#1{#1}\fi
\ifx \oauthor \undefined \def \oauthor#1{#1}\fi
\ifx \citeauthoryear \undefined \def \citeauthoryear#1{#1}\fi
\ifx \endbibitem  \undefined \def \endbibitem {}\fi
\ifx \bconflocation  \undefined \def \bconflocation#1{#1}\fi
\ifx \arxivurl  \undefined \def \arxivurl#1{\textsf{#1}}\fi
\csname PreBibitemsHook\endcsname

\bibitem[\protect\citeauthoryear{Le~M{\'e}tayer and
  De}{2016}]{lemetayer-hal-01420983}
\begin{bchapter}
\bauthor{\bsnm{Le~M{\'e}tayer}, \binits{D.}},
\bauthor{\bsnm{De}, \binits{S.J.}}:
\bctitle{{PRIAM: a Privacy Risk Analysis Methodology}}.
In: \beditor{\bsnm{Livraga}, \binits{G.}},
\beditor{\bsnm{Torra}, \binits{V.}},
\beditor{\bsnm{Aldini}, \binits{A.}},
\beditor{\bsnm{Martinelli}, \binits{F.}},
\beditor{\bsnm{Suri}, \binits{N.}} (eds.)
\bbtitle{{Data Privacy Management and Security Assurance}}.
\bpublisher{{Springer}},
\blocation{Heraklion, Greece}
(\byear{2016}).
\burl{https://hal.inria.fr/hal-01420983}
\end{bchapter}
\endbibitem

\bibitem[\protect\citeauthoryear{NIST}{2021}]{nist}
\begin{botherref}
\oauthor{\bsnm{NIST}}:
NIST Privacy Framework Core
(2021).
\url{https://www.nist.gov/system/files/documents/2021/05/05/NIST-Privacy-Framework-V1.0-Core-PDF.pdf}
\end{botherref}
\endbibitem

\bibitem[\protect\citeauthoryear{McMahan et~al.}{2017}]{fed-l-0}
\begin{bchapter}
\bauthor{\bsnm{McMahan}, \binits{B.}},
\bauthor{\bsnm{Moore}, \binits{E.}},
\bauthor{\bsnm{Ramage}, \binits{D.}},
\bauthor{\bsnm{Hampson}, \binits{S.}},
\bauthor{\bsnm{Arcas}, \binits{B.A.}}:
\bctitle{Communication-efficient learning of deep networks from decentralized
  data}.
In: \bbtitle{Artificial Intelligence and Statistics},
pp. \bfpage{1273}--\blpage{1282}
(\byear{2017}).
\bcomment{PMLR}
\end{bchapter}
\endbibitem

\bibitem[\protect\citeauthoryear{Ghosh et~al.}{2020}]{ghosh}
\begin{barticle}
\bauthor{\bsnm{Ghosh}, \binits{A.}},
\bauthor{\bsnm{Chung}, \binits{J.}},
\bauthor{\bsnm{Yin}, \binits{D.}},
\bauthor{\bsnm{Ramchandran}, \binits{K.}}:
\batitle{An efficient framework for clustered federated learning}.
\bjtitle{Advances in Neural Information Processing Systems}
\bvolume{33},
\bfpage{19586}--\blpage{19597}
(\byear{2020})
\end{barticle}
\endbibitem

\bibitem[\protect\citeauthoryear{Mansour et~al.}{2020}]{mansour}
\begin{botherref}
\oauthor{\bsnm{Mansour}, \binits{Y.}},
\oauthor{\bsnm{Mohri}, \binits{M.}},
\oauthor{\bsnm{Ro}, \binits{J.}},
\oauthor{\bsnm{Suresh}, \binits{A.T.}}:
Three approaches for personalization with applications to federated learning.
arXiv preprint arXiv:2002.10619
(2020)
\end{botherref}
\endbibitem

\bibitem[\protect\citeauthoryear{Sattler et~al.}{2020}]{sattler}
\begin{barticle}
\bauthor{\bsnm{Sattler}, \binits{F.}},
\bauthor{\bsnm{M{\"u}ller}, \binits{K.-R.}},
\bauthor{\bsnm{Samek}, \binits{W.}}:
\batitle{Clustered federated learning: Model-agnostic distributed multitask
  optimization under privacy constraints}.
\bjtitle{IEEE transactions on neural networks and learning systems}
\bvolume{32}(\bissue{8}),
\bfpage{3710}--\blpage{3722}
(\byear{2020})
\end{barticle}
\endbibitem

\bibitem[\protect\citeauthoryear{Hitaj et~al.}{2017}]{hitaj2017deep}
\begin{bchapter}
\bauthor{\bsnm{Hitaj}, \binits{B.}},
\bauthor{\bsnm{Ateniese}, \binits{G.}},
\bauthor{\bsnm{Perez-Cruz}, \binits{F.}}:
\bctitle{Deep models under the gan: information leakage from collaborative deep
  learning}.
In: \bbtitle{Proceedings of the 2017 ACM SIGSAC Conference on Computer and
  Communications Security},
pp. \bfpage{603}--\blpage{618}
(\byear{2017})
\end{bchapter}
\endbibitem

\bibitem[\protect\citeauthoryear{Nasr et~al.}{2019}]{nasr2019comprehensive}
\begin{bchapter}
\bauthor{\bsnm{Nasr}, \binits{M.}},
\bauthor{\bsnm{Shokri}, \binits{R.}},
\bauthor{\bsnm{Houmansadr}, \binits{A.}}:
\bctitle{Comprehensive privacy analysis of deep learning: Passive and active
  white-box inference attacks against centralized and federated learning}.
In: \bbtitle{2019 IEEE Symposium on Security and Privacy (SP)},
pp. \bfpage{739}--\blpage{753}
(\byear{2019}).
\bcomment{IEEE}
\end{bchapter}
\endbibitem

\bibitem[\protect\citeauthoryear{Zhu et~al.}{2019}]{deep-leakage}
\begin{botherref}
\oauthor{\bsnm{Zhu}, \binits{L.}},
\oauthor{\bsnm{Liu}, \binits{Z.}},
\oauthor{\bsnm{Han}, \binits{S.}}:
Deep leakage from gradients.
Advances in Neural Information Processing Systems
\textbf{32}
(2019)
\end{botherref}
\endbibitem

\bibitem[\protect\citeauthoryear{Dwork et~al.}{2006a}]{DworkDP1}
\begin{bchapter}
\bauthor{\bsnm{Dwork}, \binits{C.}},
\bauthor{\bsnm{McSherry}, \binits{F.}},
\bauthor{\bsnm{Nissim}, \binits{K.}},
\bauthor{\bsnm{Smith}, \binits{A.}}:
\bctitle{Calibrating noise to sensitivity in private data analysis}.
In: \beditor{\bsnm{Halevi}, \binits{S.}},
\beditor{\bsnm{Rabin}, \binits{T.}} (eds.)
\bbtitle{Theory of Cryptography},
pp. \bfpage{265}--\blpage{284}.
\bpublisher{Springer},
\blocation{Berlin, Heidelberg}
(\byear{2006})
\end{bchapter}
\endbibitem

\bibitem[\protect\citeauthoryear{Dwork et~al.}{2006b}]{DworkDP2}
\begin{bchapter}
\bauthor{\bsnm{Dwork}, \binits{C.}},
\bauthor{\bsnm{Kenthapadi}, \binits{K.}},
\bauthor{\bsnm{McSherry}, \binits{F.}},
\bauthor{\bsnm{Mironov}, \binits{I.}},
\bauthor{\bsnm{Naor}, \binits{M.}}:
\bctitle{Our data, ourselves: Privacy via distributed noise generation}.
In: \beditor{\bsnm{Vaudenay}, \binits{S.}} (ed.)
\bbtitle{Advances in Cryptology - EUROCRYPT 2006},
pp. \bfpage{486}--\blpage{503}.
\bpublisher{Springer},
\blocation{Berlin, Heidelberg}
(\byear{2006})
\end{bchapter}
\endbibitem

\bibitem[\protect\citeauthoryear{Andrew et~al.}{2021}]{adaptive}
\begin{botherref}
\oauthor{\bsnm{Andrew}, \binits{G.}},
\oauthor{\bsnm{Thakkar}, \binits{O.}},
\oauthor{\bsnm{McMahan}, \binits{B.}},
\oauthor{\bsnm{Ramaswamy}, \binits{S.}}:
Differentially private learning with adaptive clipping.
Advances in Neural Information Processing Systems
\textbf{34}
(2021)
\end{botherref}
\endbibitem

\bibitem[\protect\citeauthoryear{McMahan et~al.}{2018}]{brendan2018learning}
\begin{bchapter}
\bauthor{\bsnm{McMahan}, \binits{H.B.}},
\bauthor{\bsnm{Ramage}, \binits{D.}},
\bauthor{\bsnm{Talwar}, \binits{K.}},
\bauthor{\bsnm{Zhang}, \binits{L.}}:
\bctitle{Learning differentially private recurrent language models}.
In: \bbtitle{International Conference on Learning Representations}
(\byear{2018}).
\burl{https://openreview.net/forum?id=BJ0hF1Z0b}
\end{bchapter}
\endbibitem

\bibitem[\protect\citeauthoryear{Truex et~al.}{2020}]{ldpfl}
\begin{bchapter}
\bauthor{\bsnm{Truex}, \binits{S.}},
\bauthor{\bsnm{Liu}, \binits{L.}},
\bauthor{\bsnm{Chow}, \binits{K.-H.}},
\bauthor{\bsnm{Gursoy}, \binits{M.E.}},
\bauthor{\bsnm{Wei}, \binits{W.}}:
\bctitle{Ldp-fed: Federated learning with local differential privacy}.
In: \bbtitle{Proceedings of the Third ACM International Workshop on Edge
  Systems, Analytics and Networking},
pp. \bfpage{61}--\blpage{66}
(\byear{2020})
\end{bchapter}
\endbibitem

\bibitem[\protect\citeauthoryear{Zhao et~al.}{2020}]{zhao2020local}
\begin{barticle}
\bauthor{\bsnm{Zhao}, \binits{Y.}},
\bauthor{\bsnm{Zhao}, \binits{J.}},
\bauthor{\bsnm{Yang}, \binits{M.}},
\bauthor{\bsnm{Wang}, \binits{T.}},
\bauthor{\bsnm{Wang}, \binits{N.}},
\bauthor{\bsnm{Lyu}, \binits{L.}},
\bauthor{\bsnm{Niyato}, \binits{D.}},
\bauthor{\bsnm{Lam}, \binits{K.-Y.}}:
\batitle{Local differential privacy-based federated learning for internet of
  things}.
\bjtitle{IEEE Internet of Things Journal}
\bvolume{8}(\bissue{11}),
\bfpage{8836}--\blpage{8853}
(\byear{2020})
\end{barticle}
\endbibitem

\bibitem[\protect\citeauthoryear{Chatzikokolakis et~al.}{2013}]{broadening}
\begin{bchapter}
\bauthor{\bsnm{Chatzikokolakis}, \binits{K.}},
\bauthor{\bsnm{Andr{\'e}s}, \binits{M.E.}},
\bauthor{\bsnm{Bordenabe}, \binits{N.E.}},
\bauthor{\bsnm{Palamidessi}, \binits{C.}}:
\bctitle{Broadening the scope of differential privacy using metrics}.
In: \bbtitle{International Symposium on Privacy Enhancing Technologies
  Symposium},
pp. \bfpage{82}--\blpage{102}
(\byear{2013}).
\bcomment{Springer}
\end{bchapter}
\endbibitem

\bibitem[\protect\citeauthoryear{Biswas and
  Palamidessi}{2023}]{biswas2023privic}
\begin{botherref}
\oauthor{\bsnm{Biswas}, \binits{S.}},
\oauthor{\bsnm{Palamidessi}, \binits{C.}}:
PRIVIC: A privacy-preserving method for incremental collection of location data
(2023)
\end{botherref}
\endbibitem

\bibitem[\protect\citeauthoryear{Fernandes
  et~al.}{2022}]{Natasha_optimaldprivacy}
\begin{bchapter}
\bauthor{\bsnm{Fernandes}, \binits{N.}},
\bauthor{\bsnm{McIver}, \binits{A.}},
\bauthor{\bsnm{Palamidessi}, \binits{C.}},
\bauthor{\bsnm{Ding}, \binits{M.}}:
\bctitle{Universal optimality and robust utility bounds for metric differential
  privacy}.
In: \bbtitle{2022 IEEE 35th Computer Security Foundations Symposium (CSF)},
pp. \bfpage{348}--\blpage{363}
(\byear{2022}).
\doiurl{10.1109/CSF54842.2022.9919647}
\end{bchapter}
\endbibitem

\bibitem[\protect\citeauthoryear{Atmaca et~al.}{2022}]{atmaca2022privacy}
\begin{botherref}
\oauthor{\bsnm{Atmaca}, \binits{U.I.}},
\oauthor{\bsnm{Biswas}, \binits{S.}},
\oauthor{\bsnm{Maple}, \binits{C.}},
\oauthor{\bsnm{Palamidessi}, \binits{C.}}:
A privacy preserving querying mechanism with high utility for electric vehicles
(2022)
\end{botherref}
\endbibitem

\bibitem[\protect\citeauthoryear{Galli
  et~al.}{2023}]{Galli_Personalised_icissp23}
\begin{bchapter}
\bauthor{\bsnm{Galli}, \binits{F.}},
\bauthor{\bsnm{Biswas}, \binits{S.}},
\bauthor{\bsnm{Jung}, \binits{K.}},
\bauthor{\bsnm{Cucinotta}, \binits{T.}},
\bauthor{\bsnm{Palamidessi}, \binits{C.}}:
\bctitle{Group Privacy for Personalized Federated Learning}.
In: \bbtitle{Proceedings of the 9th International Conference on Information
  Systems Security and Privacy - ICISSP},
pp. \bfpage{252}--\blpage{263}
(\byear{2023}).
\doiurl{10.5220/0011885000003405} .
\bcomment{SciTePress - INSTICC}
\end{bchapter}
\endbibitem

\bibitem[\protect\citeauthoryear{Berk et~al.}{2021}]{berk2021fairness}
\begin{barticle}
\bauthor{\bsnm{Berk}, \binits{R.}},
\bauthor{\bsnm{Heidari}, \binits{H.}},
\bauthor{\bsnm{Jabbari}, \binits{S.}},
\bauthor{\bsnm{Kearns}, \binits{M.}},
\bauthor{\bsnm{Roth}, \binits{A.}}:
\batitle{Fairness in criminal justice risk assessments: The state of the art}.
\bjtitle{Sociological Methods \& Research}
\bvolume{50}(\bissue{1}),
\bfpage{3}--\blpage{44}
(\byear{2021})
\end{barticle}
\endbibitem

\bibitem[\protect\citeauthoryear{Chouldechova}{2017}]{chouldechova2017fair}
\begin{barticle}
\bauthor{\bsnm{Chouldechova}, \binits{A.}}:
\batitle{Fair prediction with disparate impact: A study of bias in recidivism
  prediction instruments}.
\bjtitle{Big data}
\bvolume{5}(\bissue{2}),
\bfpage{153}--\blpage{163}
(\byear{2017})
\end{barticle}
\endbibitem

\bibitem[\protect\citeauthoryear{Agarwal}{2022}]{agarwal2022trade}
\begin{botherref}
\oauthor{\bsnm{Agarwal}, \binits{S.}}:
Trade-Offs between fairness and privacy in machine learning. IJCAI 2021
  Workshop on AI for Social Good. 2021
(2022)
\end{botherref}
\endbibitem

\bibitem[\protect\citeauthoryear{Verma and Rubin}{2018}]{verma2018fairness}
\begin{bchapter}
\bauthor{\bsnm{Verma}, \binits{S.}},
\bauthor{\bsnm{Rubin}, \binits{J.}}:
\bctitle{Fairness definitions explained}.
In: \bbtitle{Proceedings of the International Workshop on Software Fairness},
pp. \bfpage{1}--\blpage{7}
(\byear{2018})
\end{bchapter}
\endbibitem

\bibitem[\protect\citeauthoryear{Hanna and Linden}{2009}]{hanna2009measuring}
\begin{botherref}
\oauthor{\bsnm{Hanna}, \binits{R.}},
\oauthor{\bsnm{Linden}, \binits{L.}}:
Measuring discrimination in education.
Technical report,
National Bureau of Economic Research
(2009)
\end{botherref}
\endbibitem

\bibitem[\protect\citeauthoryear{Makhlouf
  et~al.}{2021}]{makhlouf2021applicability}
\begin{barticle}
\bauthor{\bsnm{Makhlouf}, \binits{K.}},
\bauthor{\bsnm{Zhioua}, \binits{S.}},
\bauthor{\bsnm{Palamidessi}, \binits{C.}}:
\batitle{On the applicability of machine learning fairness notions}.
\bjtitle{ACM SIGKDD Explorations Newsletter}
\bvolume{23}(\bissue{1}),
\bfpage{14}--\blpage{23}
(\byear{2021})
\end{barticle}
\endbibitem

\bibitem[\protect\citeauthoryear{Dwork et~al.}{2012}]{dwork2012fairness}
\begin{bchapter}
\bauthor{\bsnm{Dwork}, \binits{C.}},
\bauthor{\bsnm{Hardt}, \binits{M.}},
\bauthor{\bsnm{Pitassi}, \binits{T.}},
\bauthor{\bsnm{Reingold}, \binits{O.}},
\bauthor{\bsnm{Zemel}, \binits{R.}}:
\bctitle{Fairness through awareness}.
In: \bbtitle{Proceedings of the 3rd Innovations in Theoretical Computer Science
  Conference},
pp. \bfpage{214}--\blpage{226}
(\byear{2012})
\end{bchapter}
\endbibitem

\bibitem[\protect\citeauthoryear{Hardt et~al.}{2016}]{hardt2016equality}
\begin{botherref}
\oauthor{\bsnm{Hardt}, \binits{M.}},
\oauthor{\bsnm{Price}, \binits{E.}},
\oauthor{\bsnm{Srebro}, \binits{N.}}:
Equality of opportunity in supervised learning.
Advances in neural information processing systems
\textbf{29}
(2016)
\end{botherref}
\endbibitem

\bibitem[\protect\citeauthoryear{Abadi et~al.}{2016}]{abadi}
\begin{bchapter}
\bauthor{\bsnm{Abadi}, \binits{M.}},
\bauthor{\bsnm{Chu}, \binits{A.}},
\bauthor{\bsnm{Goodfellow}, \binits{I.}},
\bauthor{\bsnm{McMahan}, \binits{H.B.}},
\bauthor{\bsnm{Mironov}, \binits{I.}},
\bauthor{\bsnm{Talwar}, \binits{K.}},
\bauthor{\bsnm{Zhang}, \binits{L.}}:
\bctitle{Deep learning with differential privacy}.
In: \bbtitle{Proceedings of the 2016 ACM SIGSAC Conference on Computer and
  Communications Security},
pp. \bfpage{308}--\blpage{318}
(\byear{2016})
\end{bchapter}
\endbibitem

\bibitem[\protect\citeauthoryear{Geyer et~al.}{2017}]{flcdp}
\begin{botherref}
\oauthor{\bsnm{Geyer}, \binits{R.C.}},
\oauthor{\bsnm{Klein}, \binits{T.}},
\oauthor{\bsnm{Nabi}, \binits{M.}}:
Differentially private federated learning: A client level perspective.
arXiv preprint arXiv:1712.07557
(2017)
\end{botherref}
\endbibitem

\bibitem[\protect\citeauthoryear{Bonawitz et~al.}{2016}]{secure-aggregation}
\begin{bchapter}
\bauthor{\bsnm{Bonawitz}, \binits{K.A.}},
\bauthor{\bsnm{Ivanov}, \binits{V.}},
\bauthor{\bsnm{Kreuter}, \binits{B.}},
\bauthor{\bsnm{Marcedone}, \binits{A.}},
\bauthor{\bsnm{McMahan}, \binits{H.B.}},
\bauthor{\bsnm{Patel}, \binits{S.}},
\bauthor{\bsnm{Ramage}, \binits{D.}},
\bauthor{\bsnm{Segal}, \binits{A.}},
\bauthor{\bsnm{Seth}, \binits{K.}}:
\bctitle{Practical secure aggregation for federated learning on user-held
  data}.
In: \bbtitle{NIPS Workshop on Private Multi-Party Machine Learning}
(\byear{2016}).
\burl{https://arxiv.org/abs/1611.04482}
\end{bchapter}
\endbibitem

\bibitem[\protect\citeauthoryear{Agarwal et~al.}{2018}]{cpsgd}
\begin{botherref}
\oauthor{\bsnm{Agarwal}, \binits{N.}},
\oauthor{\bsnm{Suresh}, \binits{A.T.}},
\oauthor{\bsnm{Yu}, \binits{F.X.X.}},
\oauthor{\bsnm{Kumar}, \binits{S.}},
\oauthor{\bsnm{McMahan}, \binits{B.}}:
cpsgd: Communication-efficient and differentially-private distributed sgd.
Advances in Neural Information Processing Systems
\textbf{31}
(2018)
\end{botherref}
\endbibitem

\bibitem[\protect\citeauthoryear{Hu et~al.}{2020}]{huetal}
\begin{barticle}
\bauthor{\bsnm{Hu}, \binits{R.}},
\bauthor{\bsnm{Guo}, \binits{Y.}},
\bauthor{\bsnm{Li}, \binits{H.}},
\bauthor{\bsnm{Pei}, \binits{Q.}},
\bauthor{\bsnm{Gong}, \binits{Y.}}:
\batitle{Personalized federated learning with differential privacy}.
\bjtitle{IEEE Internet of Things Journal}
\bvolume{7}(\bissue{10}),
\bfpage{9530}--\blpage{9539}
(\byear{2020})
\end{barticle}
\endbibitem

\bibitem[\protect\citeauthoryear{Bonawitz et~al.}{2016}]{bonawitz2017practical}
\begin{bchapter}
\bauthor{\bsnm{Bonawitz}, \binits{K.A.}},
\bauthor{\bsnm{Ivanov}, \binits{V.}},
\bauthor{\bsnm{Kreuter}, \binits{B.}},
\bauthor{\bsnm{Marcedone}, \binits{A.}},
\bauthor{\bsnm{McMahan}, \binits{H.B.}},
\bauthor{\bsnm{Patel}, \binits{S.}},
\bauthor{\bsnm{Ramage}, \binits{D.}},
\bauthor{\bsnm{Segal}, \binits{A.}},
\bauthor{\bsnm{Seth}, \binits{K.}}:
\bctitle{Practical secure aggregation for federated learning on user-held
  data}.
In: \bbtitle{NIPS Workshop on Private Multi-Party Machine Learning}
(\byear{2016}).
\burl{https://arxiv.org/abs/1611.04482}
\end{bchapter}
\endbibitem

\bibitem[\protect\citeauthoryear{Chhabra et~al.}{2010}]{chhabra2010analysis}
\begin{botherref}
\oauthor{\bsnm{Chhabra}, \binits{S.}},
\oauthor{\bsnm{Solihin}, \binits{Y.}},
\oauthor{\bsnm{Lal}, \binits{R.}},
\oauthor{\bsnm{Hoekstra}, \binits{M.}}:
An analysis of secure processor architectures.
Transactions on computational science VII,
101--121
(2010)
\end{botherref}
\endbibitem

\bibitem[\protect\citeauthoryear{Cucinotta
  et~al.}{2014}]{cucinotta2014confidential}
\begin{bchapter}
\bauthor{\bsnm{Cucinotta}, \binits{T.}},
\bauthor{\bsnm{Cherubini}, \binits{D.}},
\bauthor{\bsnm{Jul}, \binits{E.}}:
\bctitle{Confidential execution of cloud services.}
In: \bbtitle{CLOSER},
pp. \bfpage{616}--\blpage{621}
(\byear{2014})
\end{bchapter}
\endbibitem

\bibitem[\protect\citeauthoryear{Chhabra et~al.}{2021}]{chhabra2021overview}
\begin{barticle}
\bauthor{\bsnm{Chhabra}, \binits{A.}},
\bauthor{\bsnm{Masalkovait{\.e}}, \binits{K.}},
\bauthor{\bsnm{Mohapatra}, \binits{P.}}:
\batitle{An overview of fairness in clustering}.
\bjtitle{IEEE Access}
\bvolume{9},
\bfpage{130698}--\blpage{130720}
(\byear{2021})
\end{barticle}
\endbibitem

\bibitem[\protect\citeauthoryear{Ezzeldin et~al.}{2021}]{ezzeldin2021fairfed}
\begin{bchapter}
\bauthor{\bsnm{Ezzeldin}, \binits{Y.H.}},
\bauthor{\bsnm{Yan}, \binits{S.}},
\bauthor{\bsnm{He}, \binits{C.}},
\bauthor{\bsnm{Ferrara}, \binits{E.}},
\bauthor{\bsnm{Avestimehr}, \binits{S.}}:
\bctitle{Fairfed: Enabling group fairness in federated learning}.
In: \bbtitle{1st NeurIPS Workshop on New Frontiers in Federated Learning}
(\byear{2021}).
\burl{https://arxiv.org/abs/1611.04482}
\end{bchapter}
\endbibitem

\bibitem[\protect\citeauthoryear{Chu et~al.}{2021}]{chu2021fedfair}
\begin{botherref}
\oauthor{\bsnm{Chu}, \binits{L.}},
\oauthor{\bsnm{Wang}, \binits{L.}},
\oauthor{\bsnm{Dong}, \binits{Y.}},
\oauthor{\bsnm{Pei}, \binits{J.}},
\oauthor{\bsnm{Zhou}, \binits{Z.}},
\oauthor{\bsnm{Zhang}, \binits{Y.}}:
Fedfair: Training fair models in cross-silo federated learning.
arXiv preprint arXiv:2109.05662
(2021)
\end{botherref}
\endbibitem

\bibitem[\protect\citeauthoryear{Menon and Williamson}{2018}]{menon2018cost}
\begin{bchapter}
\bauthor{\bsnm{Menon}, \binits{A.K.}},
\bauthor{\bsnm{Williamson}, \binits{R.C.}}:
\bctitle{The cost of fairness in binary classification}.
In: \bbtitle{Conference on Fairness, Accountability and Transparency},
pp. \bfpage{107}--\blpage{118}
(\byear{2018}).
\bcomment{PMLR}
\end{bchapter}
\endbibitem

\bibitem[\protect\citeauthoryear{Wick et~al.}{2019}]{wick2019unlocking}
\begin{botherref}
\oauthor{\bsnm{Wick}, \binits{M.}},
\oauthor{\bsnm{Tristan}, \binits{J.-B.}}, et al.:
Unlocking fairness: a trade-off revisited.
Advances in neural information processing systems
\textbf{32}
(2019)
\end{botherref}
\endbibitem

\bibitem[\protect\citeauthoryear{Mehrabi et~al.}{2021}]{mehrabi2021survey}
\begin{barticle}
\bauthor{\bsnm{Mehrabi}, \binits{N.}},
\bauthor{\bsnm{Morstatter}, \binits{F.}},
\bauthor{\bsnm{Saxena}, \binits{N.}},
\bauthor{\bsnm{Lerman}, \binits{K.}},
\bauthor{\bsnm{Galstyan}, \binits{A.}}:
\batitle{A survey on bias and fairness in machine learning}.
\bjtitle{ACM Computing Surveys (CSUR)}
\bvolume{54}(\bissue{6}),
\bfpage{1}--\blpage{35}
(\byear{2021})
\end{barticle}
\endbibitem

\bibitem[\protect\citeauthoryear{Biswas and Rajan}{2021}]{biswas2021fair}
\begin{bchapter}
\bauthor{\bsnm{Biswas}, \binits{S.}},
\bauthor{\bsnm{Rajan}, \binits{H.}}:
\bctitle{Fair preprocessing: towards understanding compositional fairness of
  data transformers in machine learning pipeline}.
In: \bbtitle{Proceedings of the 29th ACM Joint Meeting on European Software
  Engineering Conference and Symposium on the Foundations of Software
  Engineering},
pp. \bfpage{981}--\blpage{993}
(\byear{2021})
\end{bchapter}
\endbibitem

\bibitem[\protect\citeauthoryear{Kamiran and Calders}{2012}]{kamiran2012data}
\begin{barticle}
\bauthor{\bsnm{Kamiran}, \binits{F.}},
\bauthor{\bsnm{Calders}, \binits{T.}}:
\batitle{Data preprocessing techniques for classification without
  discrimination}.
\bjtitle{Knowledge and information systems}
\bvolume{33}(\bissue{1}),
\bfpage{1}--\blpage{33}
(\byear{2012})
\end{barticle}
\endbibitem

\bibitem[\protect\citeauthoryear{Wan et~al.}{2023}]{wan2023processing}
\begin{barticle}
\bauthor{\bsnm{Wan}, \binits{M.}},
\bauthor{\bsnm{Zha}, \binits{D.}},
\bauthor{\bsnm{Liu}, \binits{N.}},
\bauthor{\bsnm{Zou}, \binits{N.}}:
\batitle{In-processing modeling techniques for machine learning fairness: A
  survey}.
\bjtitle{ACM Transactions on Knowledge Discovery from Data}
\bvolume{17}(\bissue{3}),
\bfpage{1}--\blpage{27}
(\byear{2023})
\end{barticle}
\endbibitem

\bibitem[\protect\citeauthoryear{Hashimoto
  et~al.}{2018}]{hashimoto2018fairness}
\begin{bchapter}
\bauthor{\bsnm{Hashimoto}, \binits{T.}},
\bauthor{\bsnm{Srivastava}, \binits{M.}},
\bauthor{\bsnm{Namkoong}, \binits{H.}},
\bauthor{\bsnm{Liang}, \binits{P.}}:
\bctitle{Fairness without demographics in repeated loss minimization}.
In: \bbtitle{International Conference on Machine Learning},
pp. \bfpage{1929}--\blpage{1938}
(\byear{2018}).
\bcomment{PMLR}
\end{bchapter}
\endbibitem

\bibitem[\protect\citeauthoryear{Petersen et~al.}{2021}]{petersen2021post}
\begin{barticle}
\bauthor{\bsnm{Petersen}, \binits{F.}},
\bauthor{\bsnm{Mukherjee}, \binits{D.}},
\bauthor{\bsnm{Sun}, \binits{Y.}},
\bauthor{\bsnm{Yurochkin}, \binits{M.}}:
\batitle{Post-processing for individual fairness}.
\bjtitle{Advances in Neural Information Processing Systems}
\bvolume{34},
\bfpage{25944}--\blpage{25955}
(\byear{2021})
\end{barticle}
\endbibitem

\bibitem[\protect\citeauthoryear{Noriega-Campero
  et~al.}{2019}]{noriega2019active}
\begin{bchapter}
\bauthor{\bsnm{Noriega-Campero}, \binits{A.}},
\bauthor{\bsnm{Bakker}, \binits{M.A.}},
\bauthor{\bsnm{Garcia-Bulle}, \binits{B.}},
\bauthor{\bsnm{Pentland}, \binits{A.}}:
\bctitle{Active fairness in algorithmic decision making}.
In: \bbtitle{Proceedings of the 2019 AAAI/ACM Conference on AI, Ethics, and
  Society},
pp. \bfpage{77}--\blpage{83}
(\byear{2019})
\end{bchapter}
\endbibitem

\bibitem[\protect\citeauthoryear{Cummings
  et~al.}{2019}]{cummings2019compatibility}
\begin{bchapter}
\bauthor{\bsnm{Cummings}, \binits{R.}},
\bauthor{\bsnm{Gupta}, \binits{V.}},
\bauthor{\bsnm{Kimpara}, \binits{D.}},
\bauthor{\bsnm{Morgenstern}, \binits{J.}}:
\bctitle{On the compatibility of privacy and fairness}.
In: \bbtitle{Adjunct Publication of the 27th Conference on User Modeling,
  Adaptation and Personalization},
pp. \bfpage{309}--\blpage{315}
(\byear{2019})
\end{bchapter}
\endbibitem

\bibitem[\protect\citeauthoryear{Andr{\'e}s et~al.}{2013}]{geo}
\begin{bchapter}
\bauthor{\bsnm{Andr{\'e}s}, \binits{M.E.}},
\bauthor{\bsnm{Bordenabe}, \binits{N.E.}},
\bauthor{\bsnm{Chatzikokolakis}, \binits{K.}},
\bauthor{\bsnm{Palamidessi}, \binits{C.}}:
\bctitle{Geo-indistinguishability: Differential privacy for location-based
  systems}.
In: \bbtitle{Proceedings of the 2013 ACM SIGSAC Conference on Computer \&
  Communications Security},
pp. \bfpage{901}--\blpage{914}
(\byear{2013})
\end{bchapter}
\endbibitem

\bibitem[\protect\citeauthoryear{Konečný et~al.}{2016}]{fed-l-2}
\begin{bchapter}
\bauthor{\bsnm{Konečný}, \binits{J.}},
\bauthor{\bsnm{McMahan}, \binits{H.B.}},
\bauthor{\bsnm{Yu}, \binits{F.X.}},
\bauthor{\bsnm{Richtarik}, \binits{P.}},
\bauthor{\bsnm{Suresh}, \binits{A.T.}},
\bauthor{\bsnm{Bacon}, \binits{D.}}:
\bctitle{Federated learning: Strategies for improving communication
  efficiency}.
In: \bbtitle{NIPS Workshop on Private Multi-Party Machine Learning}
(\byear{2016}).
\burl{https://arxiv.org/abs/1610.05492}
\end{bchapter}
\endbibitem

\bibitem[\protect\citeauthoryear{CMMS}{2021}]{CMMS}
\begin{botherref}
\oauthor{\bsnm{CMMS}}:
Centers for Medicare and Medicaid Services.
Accessed: 2022-09-21
(2021).
\url{https://www.cms.gov/mmrr/News/mmrr-news-2013-03-hosp-chg-data.html}
\end{botherref}
\endbibitem

\bibitem[\protect\citeauthoryear{Caldas et~al.}{2019}]{caldas2018leaf}
\begin{botherref}
\oauthor{\bsnm{Caldas}, \binits{S.}},
\oauthor{\bsnm{Duddu}, \binits{S.M.K.}},
\oauthor{\bsnm{Wu}, \binits{P.}},
\oauthor{\bsnm{Li}, \binits{T.}},
\oauthor{\bsnm{Kone{\v{c}}n{\`y}}, \binits{J.}},
\oauthor{\bsnm{McMahan}, \binits{H.B.}},
\oauthor{\bsnm{Smith}, \binits{V.}},
\oauthor{\bsnm{Talwalkar}, \binits{A.}}:
Leaf: A benchmark for federated settings.
Workshop on Federated Learning for Data Privacy and Confidentiality
(2019)
\end{botherref}
\endbibitem

\bibitem[\protect\citeauthoryear{Bassily et~al.}{2017}]{ldp-bounds}
\begin{bchapter}
\bauthor{\bsnm{Bassily}, \binits{R.}},
\bauthor{\bsnm{Nissim}, \binits{K.}},
\bauthor{\bsnm{Stemmer}, \binits{U.}},
\bauthor{\bsnm{Guha~Thakurta}, \binits{A.}}:
\bctitle{Practical locally private heavy hitters}.
In: \beditor{\bsnm{Guyon}, \binits{I.}},
\beditor{\bsnm{Luxburg}, \binits{U.V.}},
\beditor{\bsnm{Bengio}, \binits{S.}},
\beditor{\bsnm{Wallach}, \binits{H.}},
\beditor{\bsnm{Fergus}, \binits{R.}},
\beditor{\bsnm{Vishwanathan}, \binits{S.}},
\beditor{\bsnm{Garnett}, \binits{R.}} (eds.)
\bbtitle{Advances in Neural Information Processing Systems},
vol. \bseriesno{30}.
\bpublisher{Curran Associates, Inc.},
\blocation{Red Hook, NY, USA}
(\byear{2017}).
\burl{https://proceedings.neurips.cc/paper/2017/file/3d779cae2d46cf6a8a99a35ba4167977-Paper.pdf}
\end{bchapter}
\endbibitem

\bibitem[\protect\citeauthoryear{Bartlett et~al.}{2022}]{bartlett2022consumer}
\begin{barticle}
\bauthor{\bsnm{Bartlett}, \binits{R.}},
\bauthor{\bsnm{Morse}, \binits{A.}},
\bauthor{\bsnm{Stanton}, \binits{R.}},
\bauthor{\bsnm{Wallace}, \binits{N.}}:
\batitle{Consumer-lending discrimination in the fintech era}.
\bjtitle{Journal of Financial Economics}
\bvolume{143}(\bissue{1}),
\bfpage{30}--\blpage{56}
(\byear{2022})
\end{barticle}
\endbibitem

\end{thebibliography}

\end{document}